%% file: acl_latex.tex
\pdfoutput=1

\documentclass[11pt]{article}

\usepackage[preprint]{acl}

\usepackage{times}
\usepackage{latexsym}

\usepackage[T1]{fontenc}

\usepackage[utf8]{inputenc}

\usepackage{microtype}

\usepackage{inconsolata}

\usepackage{graphicx}
\usepackage{amsmath}
\usepackage{booktabs}
\usepackage{multirow}
\usepackage{colortbl}
\usepackage{xcolor}
\usepackage{arydshln}
\usepackage{makecell}
\usepackage{amssymb} 
\usepackage{CJKutf8} 
\usepackage{longtable}  
\usepackage{listings}
\definecolor{codegray}{rgb}{0.95,0.95,0.95}
\lstnewenvironment{longprompt}[1][]
{
\onecolumn\lstset{basicstyle={\small\it},prebreak={},postbreak=\mbox{\hspace{-2.25 em}},backgroundcolor=\color{codegray},frame=single,framerule=1pt,escapeinside={(*@}{@*)},#1}}
{}
\title{AutoLogi: Automated Generation of Logic Puzzles for Evaluating Reasoning Abilities of Large Language Models}

\author{
 \textbf{Qin Zhu\textsuperscript{1,2,3}}\thanks{\ \ \ Work done during internship at Qwen Team, Alibaba Group.},
 \textbf{Fei Huang\textsuperscript{1}}\thanks{\ \ \ Corresponding authors.},
 \textbf{Runyu Peng\textsuperscript{2,3}},
 \textbf{Keming Lu\textsuperscript{1}},
 \textbf{Bowen Yu\textsuperscript{1}},
 \textbf{Qinyuan Cheng\textsuperscript{2,3}},
 \\
 \textbf{Xipeng Qiu\textsuperscript{2,3}},
 \textbf{Xuanjing Huang \textsuperscript{2,3}},
 \textbf{Junyang Lin \textsuperscript{1}}{$^\dagger$}
\\
\\
 \textsuperscript{1}Qwen Team, Alibaba Group, \quad
 \textsuperscript{2}School of Computer Science, Fudan University,\\
 \textsuperscript{3}Shanghai Key Laboratory of Intelligent Information Processing, Fudan University.
\\
 \small{
  \href{mailto:email@domain}
   {\{zhuq22,rypeng22\}@m.fudan.edu.cn} } \quad
   \href{mailto:email@domain}{\{chengqy21,xpqiu,xjhuang\}@fudan.edu.cn} \\
\small{
    \href{mailto:email@domain}{\{feihu.hf,lukeming.lkm,yubowen.ybw,junyang.ljy\}@alibaba-inc.com}  
 } 
}

\begin{document}
\maketitle
\begin{abstract}

While logical reasoning evaluation of Large Language Models (LLMs) has attracted significant attention, existing benchmarks predominantly rely on multiple-choice formats that are vulnerable to random guessing, leading to overestimated performance and substantial performance fluctuations. To obtain more accurate assessments of models' reasoning capabilities, we propose an automated method for synthesizing open-ended logic puzzles, and use it to develop a bilingual benchmark, AutoLogi. Our approach features program-based verification and controllable difficulty levels, enabling more reliable evaluation that better distinguishes models' reasoning abilities. Extensive evaluation of eight modern LLMs shows that AutoLogi can better reflect true model capabilities, with performance scores spanning from 35\% to 73\% compared to the narrower range of 21\% to 37\% on the source multiple-choice dataset. Beyond benchmark creation, this synthesis method can generate high-quality training data by incorporating program verifiers into the rejection sampling process, enabling systematic enhancement of LLMs' reasoning capabilities across diverse datasets. 
\end{abstract}

\input{intro}
\input{related_work}
\input{method}
\input{experiment}
\input{conclusion}
\input{limitation}


\bibliography{custom}

\appendix

\include{appendix}

\end{document}

%% file: intro.tex
\section{Introduction}

Large Language Models (LLMs) have demonstrated remarkable capabilities across diverse applications~\cite{openai2024gpt4technicalreport, cluade, qwen2025qwen25technicalreport}. Among these capabilities, logical reasoning has emerged as a critical skill~\cite{luo2023towards,wei2022chain,yao2024tree,zhu2024dynamic}.
The increasing emphasis on reasoning capabilities has highlighted the pressing need for reliable evaluation methodologies.

\begin{figure}[t]
  \centering
  \includegraphics[width=1.0\linewidth]{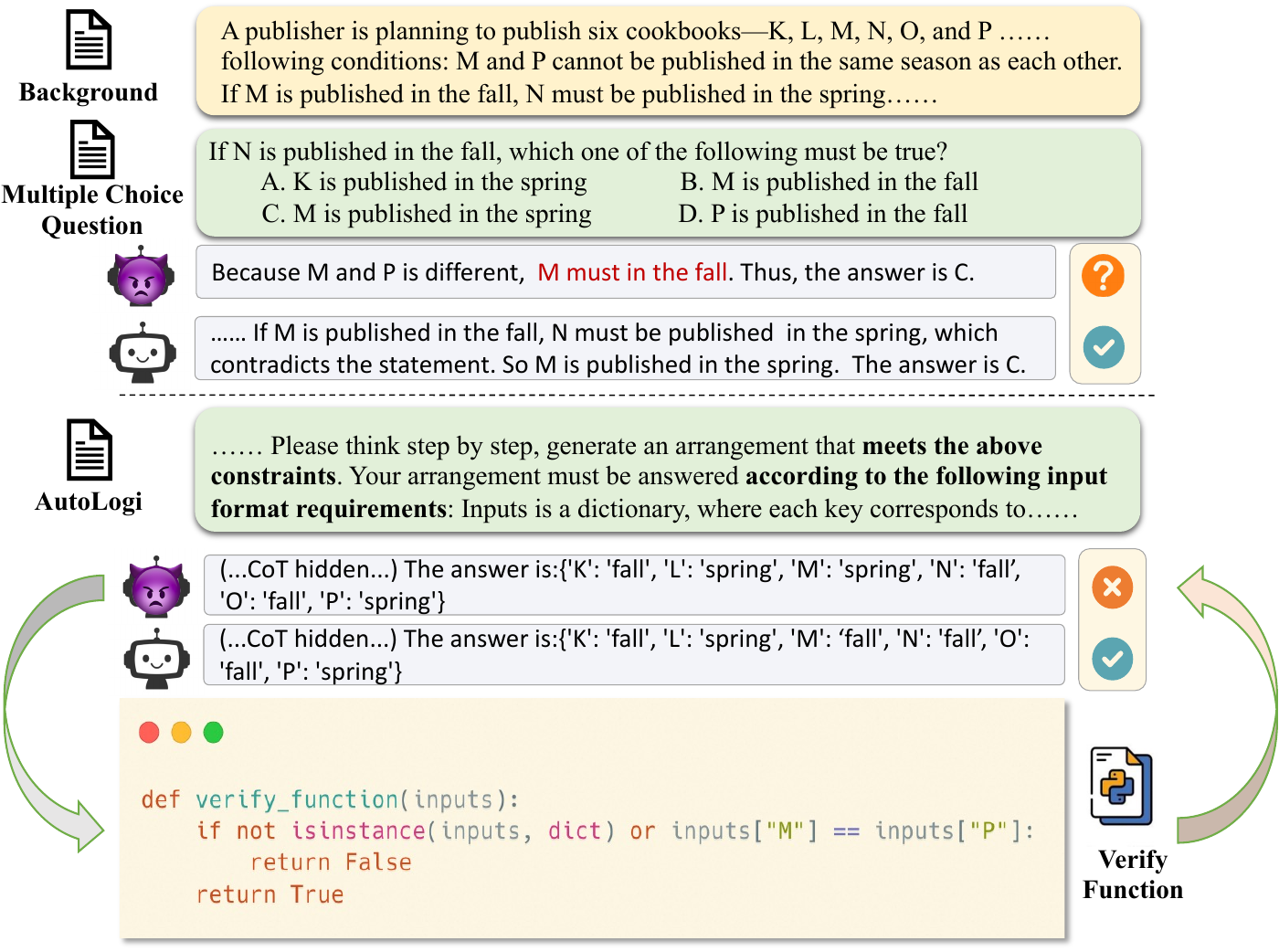}

  \caption{Comparison of evaluation processes between multiple-choice questions and our method. While multiple-choice questions may allow underperforming models to guess the correct answer, our method generates open generative questions, and utilize a verification function to validate the generated solution, providing a more accurate reflection of model performance.}
  \label{fig:intro}
\vspace{-5pt}
\end{figure}

However, the field of reasoning evaluation faces three fundamental challenges: the vulnerability to random guessing, insufficient difficulty variation to differentiate model capabilities, and high human annotation costs in dataset construction. A critical issue lies in the prevalent reliance on closed-ended questions in existing benchmarks, where models are simply required to select answers from predefined options~\cite{liu2020logiqachallengedatasetmachine,bansal2023fewshotunifiedquestionanswering,zhong2021ar,han2024folionaturallanguagereasoning,ismayilzada2023crowbenchmarkingcommonsensereasoning}. This format allows models with minimal reasoning capabilities to achieve substantially overestimated performance, leading to deeply misleading evaluation results that mask true reasoning deficiencies. The rapid advancements in LLMs are leading to the saturation of scores on existing benchmarks, reducing their effectiveness in differentiating model capabilities. Consequently, there is a growing and continuous need for the development of more challenging and distinctive benchmarks to better evaluate model performance. Nevertheless, constructing such high-quality reasoning datasets requires extensive human annotation effort, which substantially limits the scale of available evaluation resources.



To address these challenges, we introduce a novel method for automatically synthesizing open-ended logic puzzles to construct a reasoning benchmark named AutoLogi. Our approach offers three key advantages: (1) as is illustrated in Figure   ~\ref{fig:intro}, the open-ended format requires models to construct complete solutions from scratch, significantly mitigating the performance inflation caused by random guessing; (2) automated augmentation generates puzzles with varying logical constraints, enabling balanced difficulty distribution; (3) fully automated generation with minimal human verification substantially reduces benchmark construction costs. These innovations directly address the aforementioned challenges, offering a more reliable, balanced, and scalable evaluation methodology.

The proposed method follows a three-stage pipeline: information extraction from corpora for open-ended question synthesis, program-based verifier generation using advanced LLMs, and dataset augmentation for difficulty balance. Since most generation steps are based on LLMs thus potentially introducing errors, we employ a cross-validation framework where verification functions and traversal functions verify each other. The traversal functions validate puzzle correctness through exhaustive search, while verification functions check the traversal results. Although this provides only necessary conditions, our experiments show it corrects 23\% of erroneous data. Building upon this foundation, We further develop a data augmentation method that creates puzzle variants by modifying logical constraints to control difficulty levels, enabling us to expand our dataset to 1,575 English and 883 Chinese puzzles.

Based on our insights that code execution provides verifiable reward signals for reasoning tasks, we explore its utility in synthesizing high-quality training data. Through rejection-sampling with verifiers, we generate both verified correct answers for supervised fine-tuning (SFT) and correct-incorrect answer pairs for Direct Preference Optimization (DPO, \citealp{rafailov2024directpreferenceoptimizationlanguage}). This verification mechanism provides stronger guarantees of solution correctness, thereby enabling the collection of higher-quality supervised training data.

Through comprehensive experiments with eight state-of-the-art language models (including GPT-4, Claude, Qwen, and LLaMA), we evaluate our method from two aspects. First, as a benchmark, AutoLogi exhibits superior discriminative power with a wider score distribution (35.25 to 72.61) compared to traditional multiple-choice formats (21.04 to 37.39), better reflecting models' true reasoning capabilities. Second, when used for training, our synthesized dataset leads to substantial improvements on independent reasoning benchmarks, notably improving Qwen's performance on LiveBench from 30\% to 35\% at 7B scale and from 46\% to 52\% at 72B scale.

Our core contributions are:
\begin{itemize}
\item We develop a method to generate open-ended logic puzzles with controllable complexity, mitigating the performance inflation caused by random guessing and ensuring reliable evaluation through code-based verification.
\item We introduce AutoLogi, a bilingual logical reasoning benchmark that better reflects models' reasoning capabilities compared to multiple-choice formats.
\item We leverage our method to generate training data that improves model performance across multiple independent benchmarks~\footnote{We have released our training and testing datasets, along with the implementation of data synthesis methods and evaluation metrics at \url{https://github.com/8188zq/AutoLogi}.}.
\end{itemize}

%% file: related_work.tex
\section{Related Work}

\subsection{Logical Reasoning Benchmarks}


\paragraph{Question Types}
Existing logical reasoning datasets primarily use multiple-choice questions \cite{liu2020logiqachallengedatasetmachine,bansal2023fewshotunifiedquestionanswering,zhong2021ar,srivastava2023beyond,suzgun2022challenging}, true/false questions \cite{han2024folionaturallanguagereasoning,ismayilzada2023crowbenchmarkingcommonsensereasoning}, or classification tasks \cite{sinha-etal-2019-clutrr,tian2021diagnosing}. These formats, while enabling simple evaluation through keyword matching, are vulnerable to random guessing. Open-ended questions provide more rigorous assessment by requiring complete solution generation. While datasets like Zebra Puzzle \cite{prosser1993hybrid} have explored this format, our AutoLogi dataset leverages programmatic verification for more reliable evaluation of logical reasoning abilities.

\paragraph{Dataset Construction Approach}
The current mainstream approaches to constructing logical reasoning datasets are threefold: 
human annotation~\cite{clark2020transformerssoftreasonerslanguage,ismayilzada2023crowbenchmarkingcommonsensereasoning,han2024folionaturallanguagereasoning}, extraction from academic exams~\cite{liu2020logiqachallengedatasetmachine,bansal2023fewshotunifiedquestionanswering,zhong2021ar}, and synthesize~\cite{saparov2022language, sinha-etal-2019-clutrr, white2024livebench}.
Among the three methods, synthesize is most cost-effective but usually limited in diversity since they heavily rely on rule-based templates.




Our AutoLogi dataset is synthesized, but different from previous methods, we employ LLMs to reformulate questions from established academic reasoning assessments (such as LogiQA~\cite{liu2020logiqachallengedatasetmachine} and AR-LSAT~\cite{zhong2021ar}, which are derived from real human logical reasoning tests), thus ensuring rich linguistic quality and logical, real-world connected contexts. We design our synthesis process to explicitly control difficulty levels (discussed in~\ref{sec:Data_Aug}), allowing for better discrimination of model reasoning abilities. 

\subsection{Program-aided Methods}
LLMs, despite their capabilities, can generate factually incorrect responses. Integrating reliable feedback mechanisms, such as code interpreters, offers a potential solution. This approach, termed Program-aided Language model or Program of Thought, has been shown by \citet{gao2023palprogramaidedlanguagemodels} and \citet{chen2023programthoughtspromptingdisentangling} to outperform Chain of Thought methods across various domains, particularly in tasks requiring definitive answers \cite{wang2024mintevaluatingllmsmultiturn, wang2024letilearninggeneratetextual, lu2023chameleonplugandplaycompositionalreasoning}. Inspired by \cite{dong2024selfplayexecutionfeedbackimproving}'s use of program-based verifiers in training, we combine this approach with rejection sampling to generate our training data.

%% file: method.tex
\section{Method}
\begin{figure*}[t!]
    \centering
    \includegraphics[width=\linewidth]{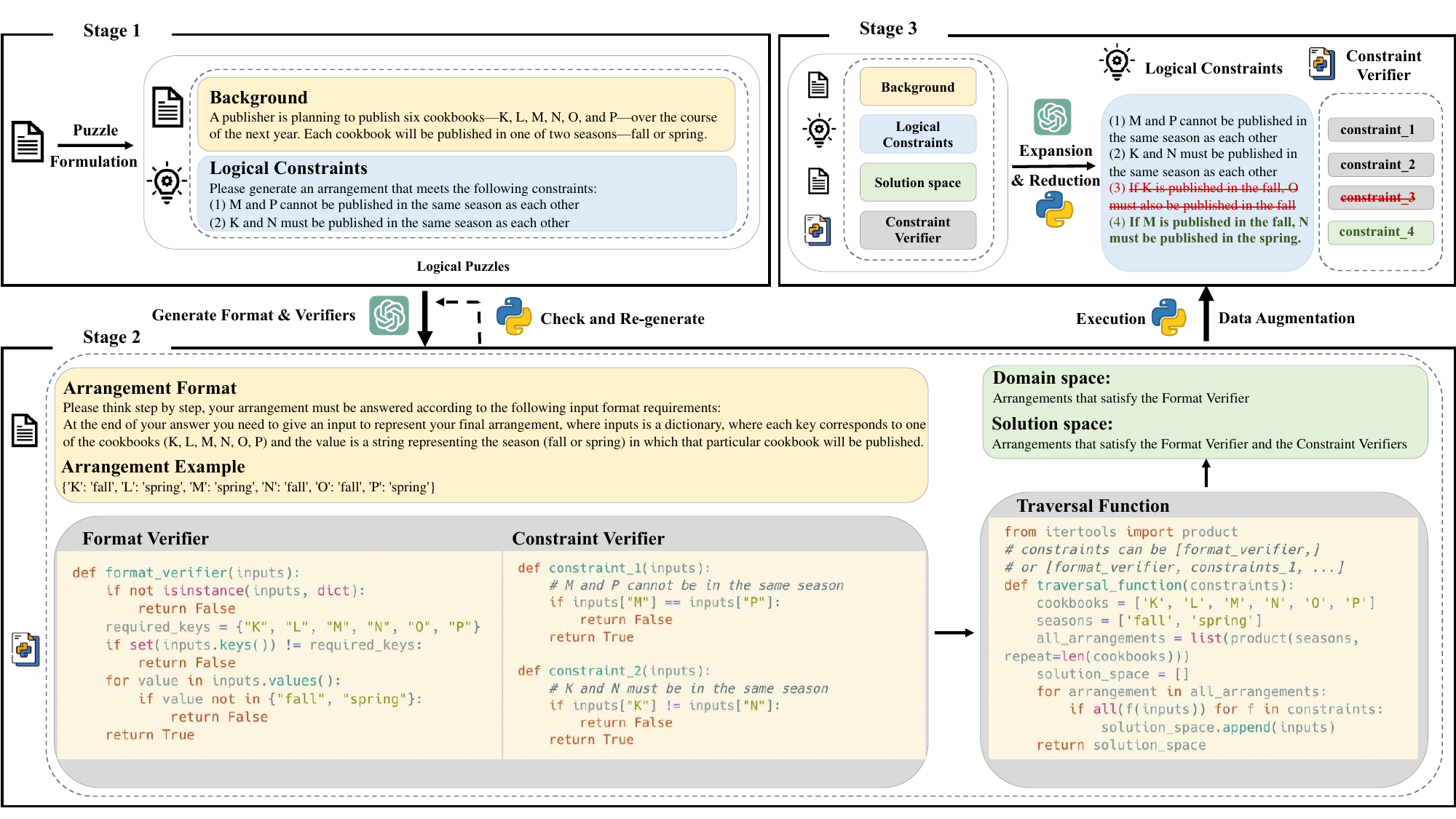}
    \caption{An overview of our method. The process consists of three stages: \textbf{Stage 1} formulates logic puzzles by extracting background information and constraints from a source corpus. \textbf{Stage 2} uses large language models (LLMs) to generate verifiers, which are programs that check puzzle solutions and ensure correct formatting. \textbf{Stage 3} augments the puzzles by adding or removing constraints to create varying difficulty levels. All three stages leverage powerful LLMs, such as GPT-4, for generation. 
   }
    \label{fig:pipeline_overview}
\end{figure*}

As shown in Figure~\ref{fig:pipeline_overview}, our method consists of three stages:  \emph{Puzzle Formulation},  \emph{Format \& Verifiers Generation}, and  \emph{Data Augmentation}.






\label{sec:question_trans}
\subsection{Puzzle Formulation}

The Puzzle Formulation stage takes a source corpus containing puzzle-related content as input and leverages advanced LLMs with direct prompting (detailed prompts in Appendix~\ref{sec:appendix-prompt}) to extract and restructure the text. The source corpus should provide background information suitable for constructing logical puzzles, for example, questions from existing multiple-choice reasoning benchmarks.
The output consists of two key components: \emph{Background} and \emph{Logical Constraints}. The \textbf{Background} provides the context and basic elements of the puzzle, defining the objects to be arranged and their value ranges. The \textbf{Logical Constraints} specify the conditions limiting valid arrangements. These components, combined with a request for test LLMs to generate any valid arrangement satisfying these constraints, as illustrated in Figure~\ref{fig:intro}, form the puzzle stem.

A notable strength of this transformation approach lies in its broad applicability - as long as the source text contains background information related to logical reasoning, it can be converted into our open-ended format, regardless of question types or the presence of answers.

\subsection{Format \& Verifiers Generation}

Stage 2 focuses on establishing a reliable evaluation mechanism. We design a program-based verification scheme consisting of three tightly coupled components: \textbf{Format Requirement} specifies the expected JSON output structure (including \textbf{Arrangement Format} and \textbf{Arrangement Example}); \textbf{Verifiers} contain program-based format verifier and constraint verifier for checking format compliance and logical constraints; \textbf{Traversal Function} searches for all possible valid solutions. This code execution-based verification approach effectively overcomes limitations of traditional evaluation methods:
since our logical puzzles may have multiple valid answers (which can grow exponentially when logical constraints are sparse), rule-based matching becomes impractical; meanwhile, model-based scoring (LLM-as-a-judge) introduces instability in complex tasks (discussed in Section \ref{sec:human-alignment}).

Given the strong correlation between Format Requirement and Verifiers, we generate these two components simultaneously using advanced LLM, with Background and Logical Constraints from Stage 1 as input. Subsequently, we generate the Traversal Function based on the produced Format Requirement and Verifiers.

To ensure the reliability of the evaluation mechanism, we propose a cross-validation method to check the correctness of LLM-generated Verifiers and Traversal Function.
We run the Traversal Function to examines all possible combinations within the value ranges defined in Background (\textbf{Domain space}), and validates them through Verifiers to determine the existence of solutions satisfying all constraints (\textbf{Solution space}). 
When no solution is found (empty solution space) or syntax errors occur, a regeneration is triggered. Through the verification, we ensure that each problem has at least one valid solution, which is a necessary condition for problem validity. The effectiveness of Traversal Function are discussed in Section \ref{sec:solution-exist}. 

\label{sec:Data_Aug}
\subsection{Data Augmentation}
Stage 3 performs data augmentation through two complementary techniques, Reduction and Expansion, to construct a dataset with balanced difficulty distribution, enabling better discrimination of models' logical reasoning capabilities.

In the \textbf{Reduction} process, we randomly select and remove one logical constraint along with its corresponding components in the Verifier. By reducing the number of constraints, the problem is simplified, yielding more problems of lower difficulty. For each problem, we iteratively remove logical constraints to generate new problems until only one constraint remains.

In the \textbf{Expansion} process, we utilize advanced LLMs to generate additional constraints and their corresponding Verifiers, taking as input the Background and Logical Constraints from Stage 1, along with the solution space and Verifier from Stage 2. To ensure data quality, we leverage the Traversal Function developed in Stage 2 to verify the solvability of newly generated problems. The expansion process terminates when either the maximum number of attempts is reached or the solution space size reduces to one.

\subsection{Synthesizing Training Data}
Beyond benchmark construction, our data synthesis method can also be used to generate model training data through rejection-sampling with a verifier, naturally obtaining two categories: verified correct responses for Supervised Fine-Tuning (SFT) and pairs of correct-incorrect responses for DPO.

A significant advantage of this approach is that rejection sampling with program-based verifiers produces more accurate training data. Unlike multiple-choice questions that risk accepting responses where models guess correctly despite flawed reasoning, potentially introducing noise into the training data, our method ensures accuracy by using program-based verifier to examine model responses.


%% file: experiment.tex
\definecolor{deepgreen}{RGB}{0, 70, 0}
\definecolor{backgreen}{RGB}{226, 240, 217}
\newcommand{\highg}{\cellcolor{backgreen}}


\begin{table*}[!htbp]
    \centering
    \small
    \begin{tabular}{cccccc}
    \toprule
\multirow{2}{*}{\textbf{}} & \multicolumn{2}{c}{\textbf{Source Corpus}} & \textbf{Stage 1} & \textbf{Stage 2} & \textbf{Stage3} \\ \cmidrule{2-3} 
                           & \textbf{\# Backgrounds}    & \textbf{\# Samples}   & \textbf{\# Sample}  & \textbf{\# Samples}  & \textbf{\# Samples} \\ \midrule
EN                         & 40                     & 230               & 210              & 206              & 1575            \\
ZH                         & 90                     & 123               & 147              & 139              & 883            \\ \bottomrule
\end{tabular}
    \caption{Statistics of our testing data. The AutoLogi benchmark corresponds to the sample number of stage 2, and the AutoLogi (Augumented) corresponds to stage 3.
 }
    \label{tab:statistics}
    \vskip -0.1in
\end{table*}

\section{Dataset Construction}
To construct the AutoLogi dataset, we leveraged two established logical reasoning datasets: AR-LSAT~\cite{zhong2021ar} and LogiQA~\cite{liu2020logiqachallengedatasetmachine}. Both datasets comprise multiple-choice questions designed to assess logical reasoning capabilities. The complete set of prompts utilized in our study is detailed in Section~\ref{sec:appendix-prompt}.

\paragraph{Testing Set Construction}
We applied our proposed method to construct the testing set through three stages, with detailed statistics presented in Table~\ref{tab:statistics}.

First, we employed GPT-4 to filter suitable samples from the source corpus, resulting in 230 English questions from AR-LSAT (40 distinct backgrounds) and 123 Chinese questions from LogiQA (90 backgrounds). In stage 1, we applied Puzzle Formulation to transform these multiple-choice questions into open-ended puzzles. To maximize coverage, we created separate cases by pairing each question stem with individual options. After deduplication, this yielded 210 English and 147 Chinese open-ended logic puzzles.

In stage 2, we utilized both GPT-4 and GPT-4o to generate format requirements and verifiers, as their complementary capabilities helped address model-specific limitations. 
Finally, in stage 3, we applied data augmentation techniques to expand our dataset to 1,575 English and 883 Chinese puzzles.

\paragraph{Training Set Construction}

We utilize the cleaned training sets from LogiQA and AR-LSAT as source corpora, applying our proposed synthesis method to generate 1,675 Chinese samples and 5,064 English samples as $D_{training}$. The rejection sampling process involves two models: Qwen2.5-7b-instruct and Qwen2.5-72b-instruct, each performing rejection sampling with 8 rounds per input to obtain corresponding candidate responses.

By employing our verifier to evaluate these candidate responses, we naturally derive two types of training datasets:
\begin{itemize}
    \item $D_{sft}$: A collection of responses verified as correct, serving as high-quality demonstrations for supervised fine-tuning.
    \item $D_{dpo}$: Pairs of correct and incorrect responses, forming natural positive-negative sample pairs (x, y\textsubscript{w}, y\textsubscript{l}) for DPO.
\end{itemize}

\begin{table}[!t]
    \centering
    \small
    \resizebox{\linewidth}{!}{
    \begin{tabular}{ccccc}
    \toprule
\textbf{}   & $D_{\text{training}}$ & $D_{\text{dpo}}$(7b) & $D_{\text{dpo}}$(72b) & $D_{\text{sft}}$(72b)  \\
\midrule
\textbf{EN} & 5064                  & 2877               & 2349                & 3724                 \\
\textbf{CN} & 1675                  & 901                & 621                 & 1170                  \\ \bottomrule    
\end{tabular}
}
    \caption{Statistics of our training data. 
 }
    \label{tab:statistics_training}
    \vskip -0.1in
\end{table}
Through eight rounds of rejection sampling, as presented in Table~\ref{tab:statistics_training}, we constructed two datasets: $D_{\text{sft}}$ and $D_{\text{pref}}$. $D_{\text{sft}}$ was formed by excluding examples without correct solutions, while $D_{\text{pref}}$ was further refined by removing instances lacking contrastive pairs. Given their trivial nature, single-constraint puzzles were excluded from both datasets.

\section{Experiment}
\subsection{Experimental Setup}

\paragraph{Models}
We evaluate 8 models in our experiments: Qwen2.5-7B/72B-Instruct~\cite{qwen2025qwen25technicalreport}, LLama3.1-8B/70B/405B-Instruct~\cite{dubey2024llama3herdmodels}, GPT-3.5-Turbo~\cite{ChatGPT}, GPT-4o-2024-08-06~\cite{openai2024gpt4technicalreport}, and Claude-3.5-Sonnet-20240620~\cite{cluade}.

\paragraph{Evaluation Datasets}
We conduct evaluations on two versions of our AutoLogi benchmark: the base version (AutoLogi) and its augmented version (AutoLogi Augmented). We also evaluate on the original multiple-choice datasets that served as our source corpus - AR-LSAT~\cite{zhong2021ar} and LogiQA~\cite{liu2020logiqachallengedatasetmachine}.
To analyze the reliability of our benchmark, we compare the performance patterns with two established reasoning benchmarks: MUSR~\cite{sprague2024musrtestinglimitschainofthought} and LiveBench (Reasoning, 2024-08-31)~\cite{white2024livebench}. 
For all evaluations, we report the mean and standard deviation across 5 independent runs to ensure reliable results.

\begin{table*}[!ht]
\centering
\small
\resizebox{\textwidth}{!}{
\setlength{\tabcolsep}{4pt}
\begin{tabular}{cccccccccc}
\toprule

\multicolumn{1}{c}{\multirow{2}{*}{\textbf{Model}}} & \multicolumn{2}{c}{\textbf{AutoLogi}} & \multicolumn{2}{c}{\textbf{AutoLogi(Augmented)}} & \multicolumn{1}{c}{\multirow{2}{*}{\textbf{AR-LSAT}}} & \multicolumn{1}{c}{\multirow{2}{*}{\textbf{LogiQA}}} & \multicolumn{1}{c}{\multirow{2}{*}{\textbf{MUSR}}} & \multicolumn{1}{c}{\multirow{2}{*}{\textbf{LiveBench}}} \\ \cmidrule(lr){2-3} \cmidrule(lr){4-5}
\multicolumn{1}{c}{} & \multicolumn{1}{c}{\textbf{EN}} & \multicolumn{1}{c}{\textbf{CN}} & \multicolumn{1}{c}{\textbf{EN}} & \multicolumn{1}{c}{\textbf{CN}} & \multicolumn{1}{c}{} & \multicolumn{1}{c}{} & \multicolumn{1}{c}{} & \multicolumn{1}{c}{} \\ \midrule
Qwen2.5-7b-instruct & $\text{27.28}_{\textcolor{deepgreen}{\pm\text{2.41}}}$ & $\text{30.94}_{\textcolor{deepgreen}{\pm\text{1.76}}}$ & $\text{43.64}_{\textcolor{deepgreen}{\pm\text{1.25}}}$ & $\text{42.08}_{\textcolor{deepgreen}{\pm\text{1.50}}}$ & $\text{22.70}_{\textcolor{deepgreen}{\pm\text{0.84}}}$ & $\text{34.42}_{\textcolor{deepgreen}{\pm\text{1.38}}}$ & $\text{47.14}_{\textcolor{deepgreen}{\pm\text{0.73}}}$ & $\text{30.67}^{\dagger}$ \\
Qwen2.5-72b-instruct & $\text{53.50}_{\textcolor{deepgreen}{\pm\text{0.93}}}$ & $\text{54.10}_{\textcolor{deepgreen}{\pm\text{0.28}}}$ & $\text{68.18}_{\textcolor{deepgreen}{\pm\text{0.77}}}$ & $\text{63.92}_{\textcolor{deepgreen}{\pm\text{0.56}}}$ & $\text{31.65}_{\textcolor{deepgreen}{\pm\text{1.89}}}$ & \textbf{47.92}$_{\textcolor{deepgreen}{\pm\text{1.37}}}$ & $\text{54.21}_{\textcolor{deepgreen}{\pm\text{0.41}}}$ & $\text{46.00}^{\dagger}$ \\
LLama3.1-8b-instruct & $\text{25.53}_{\textcolor{deepgreen}{\pm\text{3.64}}}$ & $\text{17.41}_{\textcolor{deepgreen}{\pm\text{3.61}}}$ & $\text{37.96}_{\textcolor{deepgreen}{\pm\text{1.41}}}$ & $\text{23.69}_{\textcolor{deepgreen}{\pm\text{1.33}}}$ & $\text{22.61}_{\textcolor{deepgreen}{\pm\text{2.47}}}$ & $\text{25.04}_{\textcolor{deepgreen}{\pm\text{1.40}}}$ & $\text{49.63}_{\textcolor{deepgreen}{\pm\text{0.58}}}$ & $\text{15.33}^{\dagger}$ \\
LLama3.1-70b-instruct & $\text{53.79}_{\textcolor{deepgreen}{\pm\text{2.92}}}$ & $\text{42.59}_{\textcolor{deepgreen}{\pm\text{2.41}}}$ & $\text{62.47}_{\textcolor{deepgreen}{\pm\text{0.96}}}$ & $\text{53.77}_{\textcolor{deepgreen}{\pm\text{0.95}}}$ & $\text{30.26}_{\textcolor{deepgreen}{\pm\text{1.36}}}$ & $\text{36.25}_{\textcolor{deepgreen}{\pm\text{2.55}}}$ & $\text{57.28}_{\textcolor{deepgreen}{\pm\text{0.76}}}$ & $\text{40.67}^{\dagger}$ \\
LLama3.1-405b-instruct & $\text{59.42}_{\textcolor{deepgreen}{\pm\text{3.04}}}$ & $\text{56.98}_{\textcolor{deepgreen}{\pm\text{1.79}}}$ & $\text{70.43}_{\textcolor{deepgreen}{\pm\text{1.39}}}$ & $\text{65.39}_{\textcolor{deepgreen}{\pm\text{1.07}}}$ & $\text{32.61}_{\textcolor{deepgreen}{\pm\text{1.94}}}$ & $\text{39.35}_{\textcolor{deepgreen}{\pm\text{1.10}}}$ & $\text{57.68}_{\textcolor{deepgreen}{\pm\text{0.67}}}$ & $\text{53.33}^{\dagger}$ \\
GPT-3.5-Turbo & $\text{18.93}_{\textcolor{deepgreen}{\pm\text{3.83}}}$ & $\text{22.30}_{\textcolor{deepgreen}{\pm\text{2.39}}}$ & $\text{35.25}_{\textcolor{deepgreen}{\pm\text{0.81}}}$ & $\text{34.47}_{\textcolor{deepgreen}{\pm\text{1.39}}}$ & $\text{21.04}_{\textcolor{deepgreen}{\pm\text{1.71}}}$ & $\text{26.34}_{\textcolor{deepgreen}{\pm\text{4.13}}}$ & $\text{48.33}_{\textcolor{deepgreen}{\pm\text{1.07}}}$ & $\text{26.67}^{\dagger}$ \\
GPT-4o-2024-08-06 & \textbf{62.04}$_{\textcolor{deepgreen}{\pm\text{1.59}}}$ & $\text{55.54}_{\textcolor{deepgreen}{\pm\text{3.78}}}$ & \textbf{72.61}$_{\textcolor{deepgreen}{\pm\text{0.76}}}$ & $\text{66.70}_{\textcolor{deepgreen}{\pm\text{1.25}}}$ & \textbf{37.39}$_{\textcolor{deepgreen}{\pm\text{0.91}}}$ & $\text{43.74}_{\textcolor{deepgreen}{\pm\text{1.30}}}$ & $\text{59.95}_{\textcolor{deepgreen}{\pm\text{1.13}}}$ & $\text{54.67}^{\dagger}$ \\
Claude-3.5-sonnet & $\text{60.97}_{\textcolor{deepgreen}{\pm\text{0.88}}}$ & \textbf{60.29}$_{\textcolor{deepgreen}{\pm\text{3.19}}}$ & $\text{72.53}_{\textcolor{deepgreen}{\pm\text{0.82}}}$ & \textbf{68.24}$_{\textcolor{deepgreen}{\pm\text{0.98}}}$ & $\text{34.35}_{\textcolor{deepgreen}{\pm\text{1.76}}}$ & 46.83$_{\textcolor{deepgreen}{\pm\text{3.24}}}$ & \textbf{60.53}$_{\textcolor{deepgreen}{\pm\text{0.63}}}$ & \textbf{58.68}$^{\dagger}$ \\ \bottomrule
\end{tabular}
}
\caption{Results on the original multiple-choice datasets (AR-LSAT, LogiQA), and our benchmarks (AutoLogi and Augmented AutoLogi). The accuracy was reported with a standard deviation (std) of 5 trials. Results marked with $^{\dagger}$ are sourced from the LiveBench leaderboard(2024-08-31).
}
\label{table_main}
\end{table*}

\begin{table*}[!ht]
\centering
\small
\begin{tabular}{lcccccc}
\toprule
\multirow{2}{*}{\textbf{Model}} & \multicolumn{2}{c}{\textbf{AutoLogi}} & \multirow{2}{*}{\textbf{AR-LSAT}} & \multirow{2}{*}{\textbf{LogiQA}} & \multirow{2}{*}{\textbf{MUSR}} & \multirow{2}{*}{\textbf{LiveBench}} \\ \cmidrule(lr){2-3}
& \textbf{EN} & \textbf{CN} & & & & \\
\midrule
\textit{Baseline} Qwen2.5-7b-instruct & $\text{43.64}$ & $\text{42.08}$ & $\text{22.70}$ & $\text{34.42}$ & $\text{47.14}$ & $\text{30.67}^{\dagger}$ \\
+\textit{Self-Alignment}  $DPO$ & \highg$\text{48.80}_{\textcolor{deepgreen}{\text{+5.16}}}$ & $\text{45.39}_{\textcolor{deepgreen}{\text{+3.31}}}$ & $\text{26.09}_{\textcolor{deepgreen}{\text{+3.39}}}$ & \highg$\text{38.05}_{\textcolor{deepgreen}{\text{+3.63}}}$ & $\text{47.57}_{\textcolor{deepgreen}{\text{+0.43}}}$ & \highg$\text{35.73}_{\textcolor{deepgreen}{\text{+5.06}}}$ \\
+\textit{Strong-to-Weak} $RFT$ & $\text{48.33}_{\textcolor{deepgreen}{\text{+4.69}}}$ & \highg$\text{47.54}_{\textcolor{deepgreen}{\text{+5.46}}}$ & \highg$\text{27.30}_{\textcolor{deepgreen}{\text{+4.60}}}$ & $\text{35.93}_{\textcolor{deepgreen}{\text{+1.51}}}$ & \highg$\text{49.15}_{\textcolor{deepgreen}{\text{+2.01}}}$ & $\text{30.07}_{\textcolor{deepgreen}{-\text{0.60}}}$ \\
\midrule
\textit{Baseline} Qwen2.5-72b-instruct & $\text{68.18}$ & $\text{63.92}$ & $\text{31.65}$ & $\text{47.92}$ & $\text{54.21}$ & $\text{46.00}^{\dagger}$ \\
+\textit{Self-Alignment} $DPO$ & \highg$\text{74.79}_{\textcolor{deepgreen}{\text{+6.61}}}$ & \highg$\text{69.54}_{\textcolor{deepgreen}{\text{+5.62}}}$ & \highg$\text{38.70}_{\textcolor{deepgreen}{\text{+7.05}}}$ & \highg$\text{48.14}_{\textcolor{deepgreen}{\text{+0.22}}}$ & \highg$\text{56.48}_{\textcolor{deepgreen}{\text{+2.27}}}$ & \highg$\text{52.13}_{\textcolor{deepgreen}{\text{+6.13}}}$ \\
\bottomrule
\end{tabular}
\caption{Performance comparison across different training settings. Results report the average of five trials, with standard deviations shown in Section~\ref{sec:appendix-training-details}. Subscripted values indicate performance gains over the baseline model, and the best accuracy for each setup is highlighted in \colorbox{backgreen}{green}. Results marked with $^{\dagger}$ are sourced from the LiveBench leaderboard(2024-08-31).}
\label{tab:training-set-results}
\end{table*}

\subsection{Benchmark Results}
We evaluate the effectiveness of our synthesis method in constructing high-quality logical reasoning benchmarks through extensive experiments across eight modern LLMs.
\paragraph{Superior Assessment Capability}
Experimental results in Table~\ref{table_main} demonstrate AutoLogi's distinct advantage in model capability assessment compared to traditional benchmarks. On conventional multiple-choice datasets (AR-LSAT, LogiQA), models achieve similar scores within a narrow range, making it difficult to distinguish their true capabilities. In contrast, both AutoLogi and Augmented AutoLogi reveal more pronounced performance gaps between models, enabling finer-grained capability assessment. For instance, while Qwen2.5-72b achieves unexpectedly high scores on LogiQA, this performance anomaly diverges from its relative standings across other benchmarks, suggesting potential evaluation bias in traditional tests. Notably, AutoLogi yields clear performance stratification among models, and this hierarchical distribution strongly correlates with established benchmarks (MUSR, LiveBench), validating its reliability as an evaluation framework.

\paragraph{Bilingual Evaluation Capability}
A key feature of AutoLogi is its provision of parallel English and Chinese evaluation benchmarks. Our analysis reveals that most models maintain comparable performance across both languages, indicating robust bilingual capabilities. However, we observe notable variations in cross-lingual performance: LLaMA-series models demonstrate significantly lower performance on Chinese tests compared to English, indicating limitations in Chinese language processing. In contrast, Claude, Qwen, and GPT3.5 exhibit more balanced performance across both languages, suggesting superior cross-lingual generalization ability. This bilingual evaluation dimension provides crucial insights into models' language-agnostic reasoning capabilities.

\subsection{Training Results}
We investigate the utility of our synthesized data in enhancing models' logical reasoning abilities through training experiments.

\paragraph{Training Settings}
We investigate two training settings. The first is \textbf{Self-Alignment}, where we use the training model itself to perform rejection sampling to generate preference data $D_{pref}$ (containing pairs of correct and incorrect answers), followed by reinforcement learning training using Direct Preference Optimization (DPO, \citealp{rafailov2024directpreferenceoptimizationlanguage}).
The second setting is \textbf{Strong-to-weak Distillation} based on RFT (Rejection sampling Fine-Tuning,  \citealp{yuan2023scalingrelationshiplearningmathematical}). In this approach, we leverage a more powerful model (Qwen2.5-72b-instruct) to perform rejection sampling to generate SFT data $D_{sft}$, which is then used for fine-tuning the target model.

\paragraph{Significant In-domain Improvements}
As shown in Table~\ref{tab:training-set-results}, our optimization methods achieve substantial improvements on the AutoLogi benchmark. Notably, the DPO method brings significant performance gains of 6.61\% and 5.62\% on English and Chinese tests respectively for Qwen2.5-72b, demonstrating the effectiveness of our optimization approach on the target tasks.

\paragraph{Broad Cross-domain Generalization}
Models trained on our synthesized data demonstrate strong generalization capability on tasks with similar problem context but entirely different problem formats. For instance, Qwen2.5-72b achieves a 7.05\% improvement on the AR-LSAT test set. The enhanced performance extends to tasks with substantially different source distributions, as evidenced by a 6.13\% performance gain on the LiveBench test set, thoroughly validating the immense potential of our proposed synthesis approach in generating high-quality training data to enhance models' general logical reasoning capabilities.

\paragraph{Relationship between Model Scale and Optimization Effects}
Larger models demonstrate more substantial improvements in performance. Across various benchmarks, calculating the mean improvements under the DPO setting reveals that 72B models achieve higher average performance gains (4.65\%) compared to their 7B counterparts (3.50\%). This suggests that models with stronger foundational capabilities can further unlock their logical reasoning potential when leveraging high-quality synthetic training data.

\subsection{Analysis}
\label{sec:analysis}

\paragraph{Augmentation Effectiveness Analysis}

\begin{figure}[!t]
  \centering
  \includegraphics[width=1.0\linewidth]{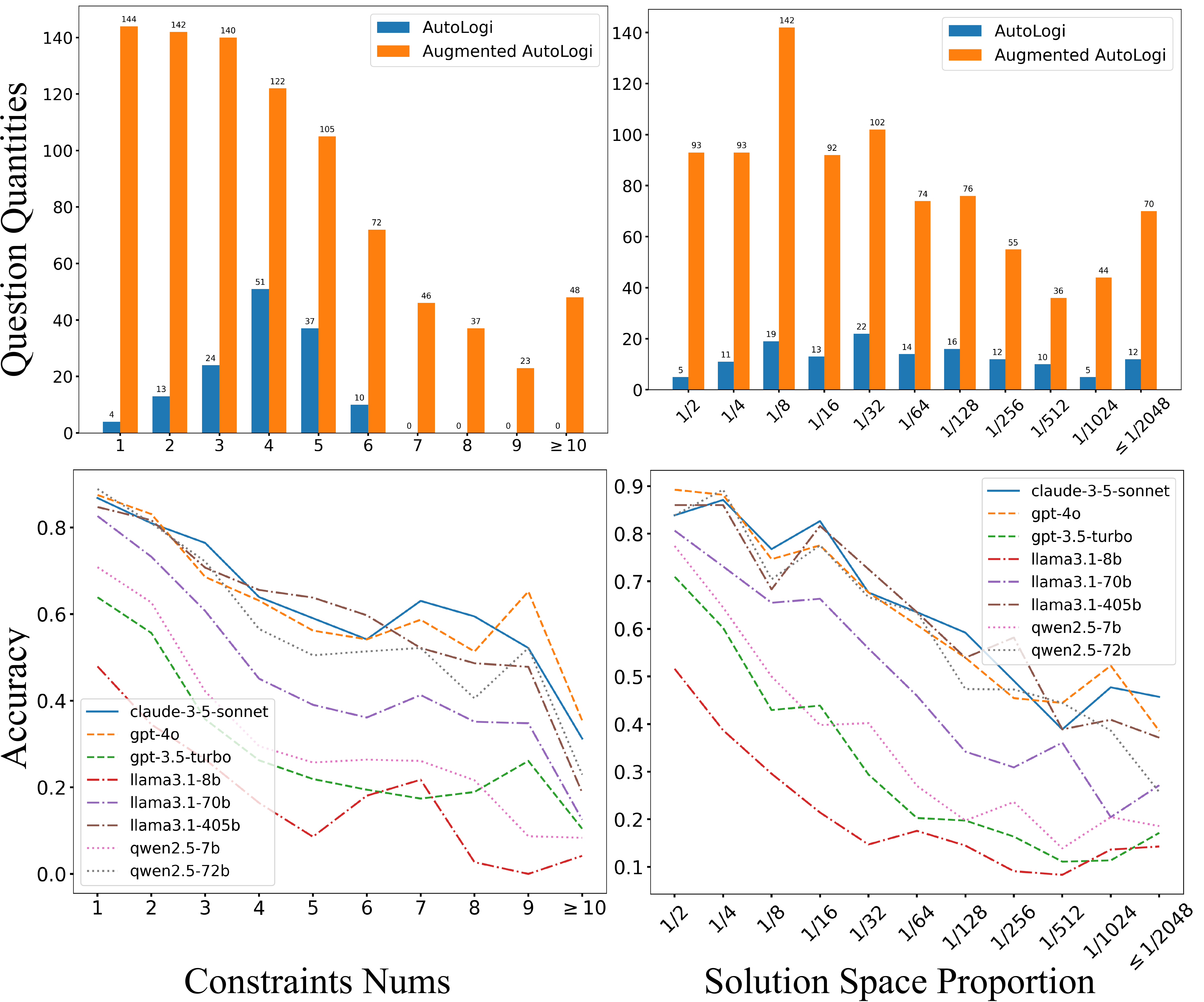}
  \caption{The question quantities on the Chinese subset of AutoLogi before and after data augmentation, and the accuracy of eight models across different constraints and solution space proportions. The figure of English subset can be found in Appendix~\ref{appendix:en}.}
  \label{fig:varying-difficulty}
\end{figure}

As illustrated in Figure~\ref{fig:varying-difficulty}, experimental results demonstrate that all models experience accuracy degradation with increasing number of constraints (the number of logical rules per puzzle) and decreasing Solution Space Ratio (the proportion of valid solutions within the total possibility space, representing theoretical random-guess probability). The consistent decline in model accuracy indicates increasing problem difficulty, and the significant correlation between these metrics validates that our augmentation strategy successfully generates problems across \textbf{diverse difficulty levels}.
The baseline Autologi dataset exhibited inherent limitations in its difficulty distribution, as problems were exclusively derived from the source corpus. This dependency resulted in an unbalanced complexity spectrum, with a maximum of 6 constraints per problem and a sparse representation of higher-difficulty instances. Our augmentation methodology addresses these limitations by introducing a more \textbf{uniform distribution} of problem complexities, thereby creating a more balanced and representative dataset for logical reasoning evaluation.

\paragraph{Cross-Validated Solution Existence Analysis}
\label{sec:solution-exist}
To ensure dataset quality, we incorporate traversal functions in Stage 2 to verify the existence of valid solution paths. This verification mechanism proves essential in detecting incorrect generated verifiers and invalid problems before they enter our dataset. For instance, in our experiments with the AutoLogi Chinese subset, among the 139 problems initially generated from 147 information units, 23\% were identified as unsolvable(11\% in English subset). Furthermore, during Stage 3's augmentation process, where additional logical constraints are introduced, the traversal function becomes indispensable as each constraint addition risks creating unsolvable problems, with approximately 30\% of LLM-generated additional constraints leading to unsolvable puzzles. Through this cross-validation method, we effectively maintain dataset integrity by eliminating invalid problems throughout our pipeline.

\begin{figure}[!t]
  \centering
  \includegraphics[width=1.0\linewidth]{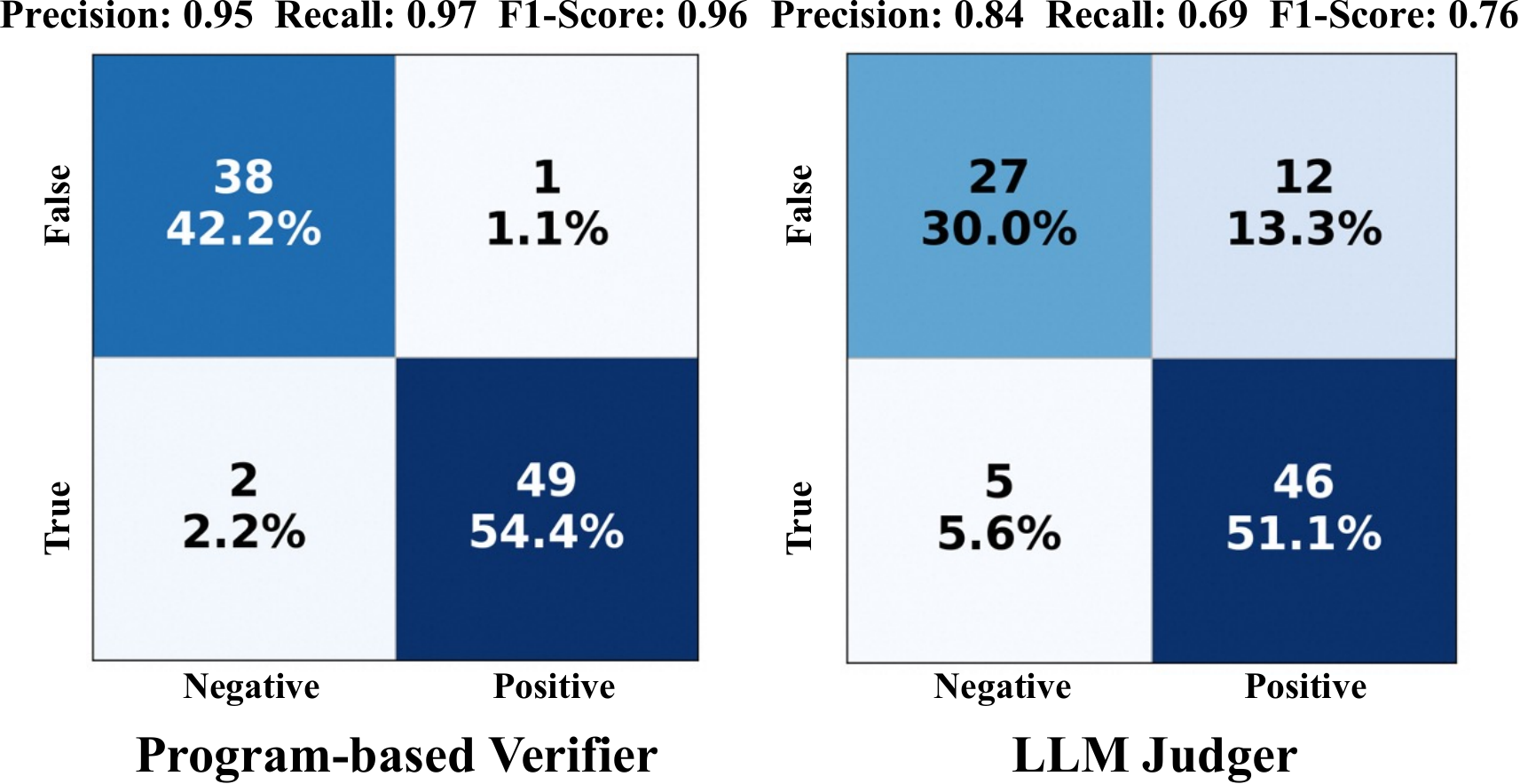}
  \caption{The precision and recall of evaluations using our verification function (Program-based Verifier) and GPT-4 as the evaluator (LLM Judger). True/False indicates the ground truth correctness of the answer. Positive/Negative is the output label predicted by Program-based Verifier or LLM Judger.}
  \label{fig:CodeVsModel}
\end{figure}

\begin{figure}[!htb]
  \centering
  \includegraphics[width=1.0\linewidth]{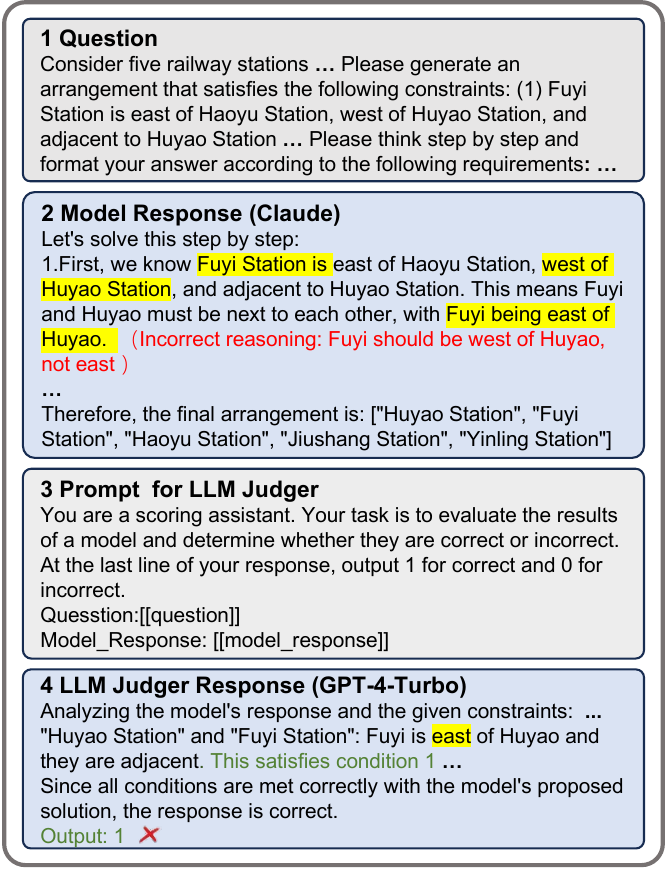}
  \caption{An example of the LLM Judger making mistakes in verifying model responses.}
  \label{fig:LLM-Judger}
\end{figure}

\paragraph{Human Alignment Analysis}
\label{sec:human-alignment}
To evaluate the effectiveness of program-based Verifier as an evaluation tool, we conducted a comparative experiment focusing on human alignment. We randomly selected 90 AutoLogi reasoning problems from diverse backgrounds and utilized Claude-3.5-Sonnet to generate responses as the model outputs to be evaluated. To ensure evaluation reliability, we collected voting results from three human evaluators as the ground truth. In our evaluation methodology, we employed both the program-based Verifier for direct answer validation and GPT-4-Turbo as an LLM judge. 
The experimental results, illustrated through the confusion matrix in Figure~\ref{fig:CodeVsModel}, demonstrate a significant performance disparity between the two evaluation methods: the program-based Verifier showed only 3 mismatches out of 90 cases (F1 score: 0.96), while the LLM judge exhibited 17 mismatches (F1 score: 0.76). Figure~\ref{fig:LLM-Judger} also presents a representative case where the LLM judge made an incorrect assessment while the program-based Verifier provided accurate evaluation.

Beyond superior accuracy, the program-based Verifier demonstrates substantial advantages in computational efficiency and cost-effectiveness, particularly crucial for large-scale evaluations. These experimental findings strongly validate the reliability and practicality of utilizing program-based Verifier as an evaluation tool.

\paragraph{Error Analysis}
Table~\ref{table_main} reveals that even advanced LLMs exhibit error rates exceeding 30\% in logical reasoning tasks. To gain a comprehensive understanding of these limitations, we conducted an in-depth error analysis using Claude-3.5-Sonnet as our representative model. Based on the 90 problems from our human alignment experiment in Section~\ref{sec:human-alignment}, we enlisted human annotators to provide detailed explanations for all failure cases. Through systematic analysis of these explanations, we categorized the error sources into four primary types: Incorrect Logical Inference (incorrect deductive steps, 81\%), Contradictory Conclusion (inconsistencies between inference and conclusions, 13\%), Unanalyzed Condition (failure to consider certain given logical constraint, 3\%), and Inconsistent Format (structural or presentational issues in responses, 3\%). Detailed statistical analysis of these error patterns and specific case studies for each category are thoroughly discussed in Appendix~\ref{sec:appendix-error-analysis} and Appendix~\ref{sec:appendix-case}, respectively.


%% file: conclusion.tex
\section{Conclusion}
In conclusion, we present a method for automatically synthesizing open-ended logic puzzles, which we use to create the bilingual benchmark AutoLogi and high-quality training data. Our approach provides more accurate assessment of LLMs' reasoning capabilities by mitigating random guessing and enabling controllable difficulty levels, while the synthesized training data proves effective in enhancing models' reasoning abilities. 

%% file: limitation.tex
\section*{Limitations}

Our methods may contains the following limitations:

\paragraph{Dependence on LLMs} Our method relies on advanced large language models (LLMs) such as GPT-4 for generating logic puzzles and verification functions. If the only accessible LLMs have limited capacity, our method may suffer from a high failure rate in synthesizing the questions. However, as LLMs continue to advance, we expect the effectiveness and reliability of our approach to improve accordingly.

\paragraph{Verification Function Limitations} Although AutoLogi uses program-based verification to handle multiple valid solutions, this approach does not guarantee that the verification functions are perfect, especially in complex conditions. In our experiment, we still find there is about 3\% of validation results contain errors (Section \ref{sec:analysis}).

%% file: appendix.tex
\definecolor{bblue}{HTML}{4F81BD}
\definecolor{rred}{HTML}{C0504D}
\definecolor{ggreen}{HTML}{9BBB59}
\definecolor{ppurple}{HTML}{9F4C7C}
\definecolor{darkGreen}{rgb}{0.2,0.5,0.2}
\definecolor{mydarkblue}{rgb}{0,0.08,0.45}
\newcommand{\checkme}[1]{\textcolor{violet}{#1}}
\newcommand{\cellcheck}[0]{\makecell[c]{\checkmark}}

\section{Appendix A}
\label{sec:appendix}

\subsection{Difficulty Control through Data Augmentation}
\label{appendix:en}
\begin{figure*}[!htb]
\centering
\includegraphics[width=1.0\linewidth]{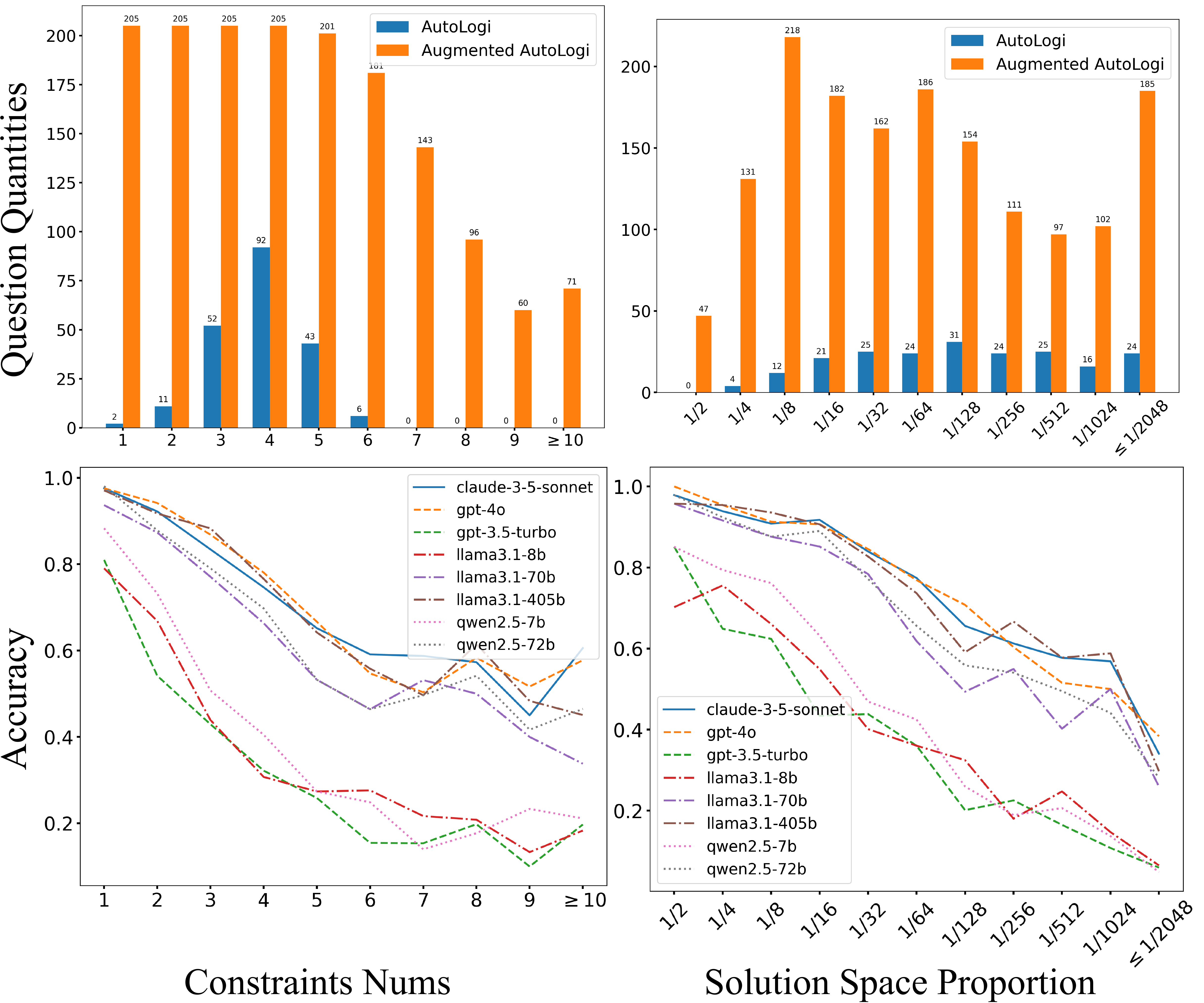}
\caption{Distribution of questions in AutoLogi's English subset before and after data augmentation, alongside the performance analysis of eight models across varying constraints and solution space proportions.}
\label{fig:varying-difficulty-en}
\end{figure*}

The experimental results on the English subset are presented in Figure~\ref{fig:varying-difficulty-en}. Our analysis reveals a clear negative correlation between model performance and two key difficulty metrics: the number of constraints and the solution space size. Specifically, model accuracy demonstrates a consistent decline as the number of constraints increases and as the solution space becomes more constrained. These findings align closely with our previous observations on the Chinese dataset, reinforcing the language-agnostic nature of these performance patterns.

The augmented dataset achieves a more balanced distribution across these difficulty indicators, enabling a more nuanced and comprehensive evaluation of model capabilities. Through systematic manipulation of constraint density and solution space, AutoLogi demonstrates its effectiveness in generating targeted evaluation scenarios with controlled difficulty levels.

\subsection{Error Analysis}
\label{sec:appendix-error-analysis}
To gain insights into improving model performance in logical reasoning tasks, we conducted a systematic analysis of incorrect responses. We examined 90 samples, one from each background category in AutoLogi's Chinese dataset, focusing on Claude 3.5 Sonnet's responses. Among these samples, the model failed on 37 cases. We categorized these errors into four distinct types:

\begin{enumerate}
\item \textbf{Unanalyzed Condition}: Cases where the model completely overlooked critical conditions in the problem statement.

\item \textbf{Incorrect Logical Inference}: Instances where the model made erroneous deductions when processing specific logical conditions.

\item \textbf{Contradictory Conclusion}: Cases where despite appearing to follow a sound reasoning process, the model's final conclusion contradicted its own analysis.

\item \textbf{Inconsistent Format}: Responses that deviated from the required output format, including issues with naming conventions and data structure consistency.
\end{enumerate}

\begin{figure}[!htb]
\centering
\includegraphics[width=0.9\linewidth]{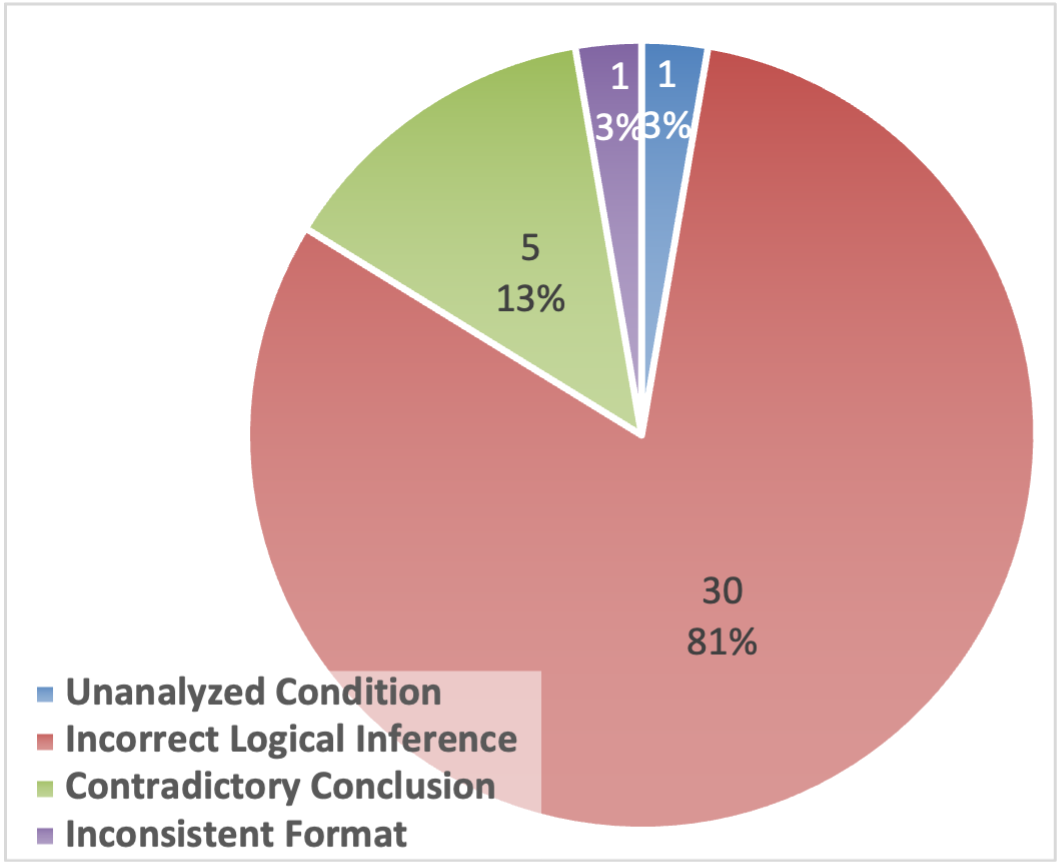}
\caption{Distribution of error types in model responses, analyzing 37 incorrect answers from Claude across 90 test questions.}
\label{fig:erro-types}
\end{figure}

Detailed examples for each error category are provided in Appendix~\ref{sec:appendix-case}. The distribution of these errors across the 37 failed cases is illustrated in Figure~\ref{fig:erro-types}. Our analysis reveals that Claude 3.5 Sonnet demonstrates strong instruction-following capabilities and processes constraints systematically. Consequently, Inconsistent Format and Unanalyzed Condition errors were relatively rare. The most prevalent issue was \textbf{Incorrect Logical Inference}.

Further investigation into the Incorrect Logical Inference cases revealed two primary challenges:

\begin{enumerate}
\item \textbf{Complex Logic}: Errors emerged when handling logical conditions that required multiple interconnected judgments.

\item \textbf{Misinterpretation of Logical Negations}: Common mistakes included misunderstanding logical operators, particularly in cases involving negations (e.g., confusing "not less than" with "less than").
\end{enumerate}

\subsection{Error Cases}
\label{sec:appendix-case}
Following the analysis of four distinct error types and their distribution in Section~\ref{sec:appendix-error-analysis}, here we present one representative example for each type from the AutoLogi Chinese dataset. These examples also serve as sample problems demonstrating how the system processes logical reasoning questions. While the AutoLogi dataset contains both Chinese and English problems, we focus on Chinese examples here since English problems typically have much longer problem descriptions. For each example, we provide both the original Chinese text and its English translation.

The following examples illustrate: Unanalyzed Condition (Table~\ref{tab:logical-reasoning-example1}), Incorrect Logical Inference (Table~\ref{tab:logical-reasoning-example2}), Contradictory Conclusion (Table~\ref{tab:logical-reasoning-example3}), and Inconsistent Format (Table~\ref{tab:logical-reasoning-example4}).

\begingroup
\begin{table*}[!ht]
\centering
\small
\begin{tabular}{p{\linewidth}}
\toprule
\textbf{\textsc{Prompt (Chinese):}} \begin{CJK*}{UTF8}{gbsn}有7名运动员参加男子5千米的决赛，他们是:S、T、U、W、X、Y和Z。运动员穿的服装不是红色，就是绿色，没有运动员同时到达终点。请你生成一个安排方案满足以下约束：\
(1) 相继到达终点的运动员，他们的服装不全是红色的。\
(2) Y在T和W之前的某一时刻到达了终点。\
(3) 在Y之前到达终点的运动员，恰好有两位穿的是红色服装。\
(4) S是第六个到达终点的运动员。\
(5) Z在U之前的某一时刻到达终点。\end{CJK*}\\
\begin{CJK*}{UTF8}{gbsn}
请你一步一步思考，你的安排方案必须按照为以下输入格式要求来进行回答：
在回答的最后，你需要给出一个输入，以表示你的最终安排方案, 其中 inputs
是一个字典，包含两个键： "order" 和 "colors"。
- inputs["order"] 是一个包含7个元素的 list，元素为运动员的名字（字符串），
表示运动员的到达顺序。所有可能的元素为["S", "T", "U", "W", "X", "Y", "Z"]。
- inputs["colors"] 是一个字典，键为运动员的名字（字符串），值为 "red" 或
"green"（字符串），表示运动员的衣服颜色。
下面提供一个输入样例（注意这只是一个合法输入的例子，不一定是正确方
案）：\end{CJK*}\\
\{
"order": ["T", "Y", "Z", "U", "W", "S", "X"],
"colors": \{S": "red", "T": "green", "U": "red", "W": "green", "X": "red",
"Y": "green", "Z": "green"\}
\}
\\
\midrule
\textbf{\textsc{Prompt (English,translated):}} There are 7 athletes participating in the men's 5-kilometer final, and they are: S, T, U, W, X, Y, and Z. The athletes' outfits are either red or green, and no athletes reached the finish line simultaneously. Please generate an arrangement that meets the following constraints:\\
(1) The athletes who reached the finish line consecutively do not all wear red outfits.\
(2) Y reached the finish line at some point before T and W.\
(3) Among the athletes who reached the finish line before Y, exactly two wore red outfits.\
(4) S is the sixth athlete to reach the finish line.\
(5) Z reached the finish line at some point before U.\\
Please think step by step, and your arrangement must be answered according to
the following input format requirements:
At the end of your answer, you need to provide an input to represent your final
arrangement. The inputs should be a dictionary containing two keys: "order" and
"colors".
- `inputs["order"]` is a list containing 7 elements, with each element being the
name of an athlete (string) representing the order in which the athletes reached
the finish line. Possible elements are `["S", "T", "U", "W", "X", "Y", "Z"]`.
- `inputs["colors"]` is a dictionary where the key is the name of an athlete (string)
and the value is "red" or "green" (string), representing the color of the athlete's
outfit.
Below is an example input (note this is just an example of valid input, not
necessarily the correct arrangement):\\
\{
"order": ["T", "Y", "Z", "U", "W", "S", "X"],
"colors": \{S": "red", "T": "green", "U": "red", "W": "green", "X": "red",
"Y": "green", "Z": "green"\}
\}\\
\midrule
\textbf{\textsc{Model Response (Chinese):}} \begin{CJK*}{UTF8}{gbsn}好的,让我们一步步思考来满足所有约束条件:\
1. 首先,我们知道S是第六个到达终点的,所以我们可以先把S放在第六位。\
{\color{darkGreen}2. Y在T和W之前到达终点。\ {\color{blue}（这里进行了简单重复没有分析）}}
3. Z在U之前到达终点,所以Z必须在U前面。\
4. 在Y之前到达终点的运动员,恰好有两位穿的是红色服装。这意味着Y之前的运动员中,必须有两个红色,一个绿色。\
5. 相继到达终点的运动员,他们的服装不全是红色的。这意味着不能有两个连续的红色。\

    根据这些条件,我们可以尝试以下安排:\
    1. X (红色)\
    2. T (绿色)\
    3. Y (绿色)\
    4. Z (红色)\
    5. W (绿色)\
    6. S (红色)\
    7. U (绿色)\
    
    这个安排满足所有条件:\
    - S在第六位\
    - Y在T和W之前\
    - Z在U之前\
    - Y之前有两个红色(X和T)\
    - 没有连续的红色\
    - 总共有3个红色,4个绿色。
    因此,最终的输入格式如下:\end{CJK*} \\
\begin{verbatim}
{"order": ["X", "T", "Y", "Z", "W", "S", "U"],
"colors": {S": "red","T": "green","U": "green","W": "green","X": "red","Y":"green","Z": "red"}}
\end{verbatim}\\
    \midrule
\textbf{\textsc{Model Response(English,translated):}} Let's think step by step to meet all the constraints:\
1. First, we know S is the sixth to reach the finish line\
{\color{darkGreen}2. Y reached the finish line before T and W\ {\color{blue}(This was merely repeated without proper analysis)}}
3. Z reached the finish line before U\
4. Among the athletes before Y, exactly two wore red outfits\
5. No consecutive athletes can wear red\\
Based on these conditions, we can arrange as follows:\
1. X (red)\
2. T (green)\
3. Y (green)\
4. Z (red)\
5. W (green)\
6. S (red)\
7. U (green)\

Final solution in required format:\\
\begin{verbatim}
{"order": ["X", "T", "Y", "Z", "W", "S", "U"],
"colors": {"S": "red", "T": "green", "U": "green","W": "green", "X": "red", "Y": "green","Z": "red"} }
\end{verbatim}\\
\midrule
\textbf{Error Type:} \textbf{Unanalyzed Condition} \\
{\color{blue}There was no analysis of the constraint "(2) Y reached the finish line at some point before T and W".}\\
\bottomrule
\end{tabular}
\caption{Error Case: Unanalyzed Condition}
\label{tab:logical-reasoning-example1}
\end{table*}
\endgroup

\begingroup
\begin{table*}[!ht]
\centering
\small
\begin{tabular}{p{\linewidth}}
\toprule
\textbf{\textsc{Prompt (Chinese):}} \begin{CJK*}{UTF8}{gbsn}某国东部沿海有5个火山岛E、F、G、H、I，它们由北至南排成一条直线。\
请你生成一个安排方案满足以下约束：\
(1) F与H相邻并且在H的北边\
(2) I和E相邻\
(3) G在F的北边某个位置\
(4) G和E相邻\end{CJK*}\\
\begin{CJK*}{UTF8}{gbsn}
请你一步一步思考，你的安排方案必须按照为以下输入格式要求来进行回答：
在回答的最后，你需要给出一个输入，以表示你的最终安排方案, 其中 inputs
是一个包含5个元素的list，元素为字符串，取值范围是["E","F","G","H","I"]。
下面提供一个输入样例（注意这只是一个合法输入的例子，不一定是正确方
案）：
['F', 'H', 'I', 'E', 'G']
\end{CJK*}\\
\midrule
\textbf{\textsc{Prompt (English,translated):}} There are 5 volcanic islands, E, F, G, H, and I, located along the eastern coast of a country, arranged in a straight line from north to south.\
Please generate an arrangement that satisfies the following constraints:\
1. F is adjacent to H and is located north of H.\
2. I is adjacent to E.\
3. G is located somewhere north of F.\
4. G is adjacent to E.\\
Please think step by step, and your arrangement must follow the input format
requirements below:
At the end of your answer, you need to provide an input to represent your final
arrangement. The input should be a list containing 5 elements, with each element
being a string from the range `["E", "F", "G", "H", "I"]`.
Below is an example input (note that this is just an example of valid input and may
not be the correct arrangement):
['F', 'H', 'I', 'E', 'G']\\
\midrule
\textbf{\textsc{Model Response (Chinese):}} \begin{CJK*}{UTF8}{gbsn}让我们一步步思考这个问题：\
1. 根据约束(1)，F与H相邻并且在H的北边。所以我们知道F和H必须相邻，且F在H的北边。\
2. 根据约束(3)，G在F的北边某个位置。这意味着G必须在F的北边，但不一定紧邻F。\
{\color{darkGreen}3. 根据约束(4)，G和E相邻。结合约束(3)，我们可以推断E必须在G的北边，因为G已经在F的北边了。\color{blue}{(推理错误：G不一定在F北边)}}\
4. 根据约束(2)，I和E相邻。考虑到前面的推论，I只能在E的北边或者H的南边。\
5. 综合以上分析，我们可以得出一个可能的排列：G-E-I-F-H 或 I-G-E-F-H。\
6. 考虑到题目要求从北到南排列，I-G-E-F-H 是符合所有约束的一个有效排列。因此，最终的安排方案如下：\end{CJK*}\\
\begin{verbatim}
['I', 'G', 'E', 'F', 'H']
\end{verbatim}\\
\begin{CJK*}{UTF8}{gbsn}
这个排列满足所有给定的约束条件：
- F与H相邻并且在H的北边
- I和E相邻
- G在F的北边
- G和E相邻\end{CJK*}\\

\midrule
\textbf{\textsc{Model Response (English,translated):}} Let's think through this problem step by step:\
1. According to constraint (1), F is adjacent to H and is north of H. So, we know F and H must be adjacent, with F to the north of H.\
2. According to constraint (3), G is located somewhere north of F. This means G must be north of F, but not necessarily immediately adjacent.\
{\color{darkGreen}3. According to constraint (4), G is adjacent to E. Combining this with constraint (3), we can infer that E must be north of G because G is already north of F.\color{blue}{(Logical error: G is not necessarily north of F)}}\
4. According to constraint (2), I is adjacent to E. Considering the previous inferences, I can only be either north of E or south of H.\
5. Combining all the above analysis, one possible arrangement is G-E-I-F-H or I-G-E-F-H.\
6. Considering the requirement that the arrangement is from north to south, I-G-E-F-H is a valid arrangement that satisfies all constraints. 
Therefore, the final arrangement is as follows:
\begin{verbatim}
['I', 'G', 'E', 'F', 'H']
\end{verbatim}
This arrangement satisfies all the given constraints:
- F is adjacent to H and is north of H.
- I is adjacent to E.
- G is north of F.
- G is adjacent to E\\
\midrule
\textbf{Error Type:} \textbf{Incorrect Logical Inference} \\
{\color{blue}Error in step 3 of the reasoning: The conclusion that "E must be north of G" cannot be derived from the premises "G is north of F" and "G is adjacent to E"}\\
\bottomrule
\end{tabular}
\caption{Error Case: Incorrect Logical Inference}
\label{tab:logical-reasoning-example2}
\end{table*}
\endgroup

\begingroup
\begin{table*}[!ht]
\centering
\small
\begin{tabular}{p{\linewidth}}
\toprule
\textbf{\textsc{Prompt (Chinese):}} \begin{CJK*}{UTF8}{gbsn}G、H、J、K、L、M和O七人结伴出游。在他们4天的露营生活中，每天要安排两个人值班为大家做饭。要求这4天的人员安排不能重复，且恰好有一人值班两天。\
请你生成一个安排方案满足以下约束：\
(1) J要排在H值班的次日。\
(2) 值班两天的那个人必须排在第四天，不能排在第三天。\
(3) G必须与J和O中的某一个人排在同一天。\
(4) 第一天或第二天至少有一天安排K。\
(5) O必须排在第三天。\end{CJK*}\\
\begin{CJK*}{UTF8}{gbsn}
请你一步一步思考，你的安排方案必须按照为以下输入格式要求来进行回答：
在回答的最后，你需要给出一个输入，以表示你的最终安排方案,
其中 inputs是一个字典，包含四个键：`"day1"`, `"day2"`, `"day3"`和`"day4"`。
`inputs["day1"]`, `inputs["day2"]`, `inputs["day3"]`, `inputs["day4"]`都是一个包含
两个元素的 set，包含值班人员名字（字符型）。取值范围是['G', 'H', 'J', 'K', 'L',
'M', 'O']。
下面提供一个输入样例（注意这只是一个合法输入的例子，不一定是正确方
案）：
\{'day1': \{'K', 'G'\}, 'day2': \{'K', 'H'\}, 'day3': \{'O', 'L'\}, 'day4': \{'J', 'M'\}\}
\end{CJK*}\\
\midrule
\textbf{\textsc{Prompt (English,translated):}} Seven people, G, H, J, K, L, M, and O, go on a trip together. During their 4-day camping, two people are assigned each day to cook for the group. The assignment must not be repeated over the 4 days, and exactly one person must be on duty for two days.\
Please generate an arrangement that satisfies the following constraints:\
1. J must be on duty the day after H.\
2. The person who is on duty for two days must be assigned on the fourth day and cannot be assigned on the third day.\
3. G must be on the same day as either J or O.\
4. K must be scheduled on either the first or the second day.\
5. O must be scheduled on the third day.\\
Please think step by step, and your arrangement must follow the input format
requirements below:
At the end of your answer, you need to provide an input to represent your final
arrangement. The input should be a dictionary containing four keys: "day1", "day2",
"day3", and "day4".
`inputs["day1"]`, `inputs["day2"]`, `inputs["day3"]`, and `inputs["day4"]` are each
*sets* containing two elements representing the names of the people on duty
(*strings*). The range of values is `['G', 'H', 'J', 'K', 'L', 'M', 'O']`.
Below is an example input (note that this is just an example of valid input and may
not be the correct arrangement):
\{'day1': \{'K', 'G'\}, 'day2': \{'K', 'H'\}, 'day3': \{'O', 'L'\}, 'day4': \{'J', 'M'\}\} \\
\midrule
\textbf{\textsc{Model Response (Chinese):}} \begin{CJK*}{UTF8}{gbsn}好的，我会一步步思考来制定一个满足所有约束条件的安排方案。\
1. O必须排在第三天，所以先确定：\
- inputs["day3"] = {'O', 某人}\
{\color{darkGreen}2. J要排在H值班的次日，所以H必须在第1或第2天，J在第2或第3天。但J不能和O同天，所以：\
- inputs["day2"] = {'H', 某人}\
- inputs["day3"] = {'O', 'J'}  \color{blue}{(这里要求'J'安排在第三天，与最终给出的结论是相互矛盾的)}}\
3. 值班两天的人必须在第四天，不能在第三天。G必须和J或O同一天，但G不能是值班两天的人(因为G不在第四天)，所以G必须和O同一天：\
- inputs["day3"] = {'O', 'G'}\
- inputs["day4"] = {'J', 值班两天的人}\
4. 第一天或第二天至少有一天安排K。我们可以把K安排在第一天：\
- inputs["day1"] = {'K', 某人}\
5. 剩下L和M还没安排。L可以安排在第一天，M可以安排在第二天：\
- inputs["day1"] = {'K', 'L'}\
- inputs["day2"] = {'H', 'M'}\
6. 现在，值班两天的人只能是M，因为其他人都只值班了一次。
综上所述，最终的安排方案是：
\begin{verbatim}
{'day1': {'K', 'L'}, 'day2': {'H', 'M'}, 'day3': {'O', 'G'}, 'day4': {'J', 'M'}}
\end{verbatim}
这个方案满足了所有给定的约束条件。\end{CJK*}\\
\midrule
\textbf{\textsc{Model Response (English,translated):}} Let's think step by step to create an arrangement that satisfies all the constraints.\
1. O must be scheduled on the third day, so first we set:\
- inputs["day3"] = {'O', someone}\
{\color{darkGreen}2. J must be on duty the day after H, so H must be on the 1st or 2nd day, and J on the 2nd or 3rd day. But J cannot be on the same day as O, so:\
- inputs["day2"] = {'H', someone}\
- inputs["day3"] = {'O', 'J'} \color{blue}{(There's a contradiction between requiring 'J' to be on day 3 here and the final arrangement presented)}}\
3. The person on duty for two days must be on the fourth day, and not on the third day. G must be on the same day as either J or O, but G cannot be the person on duty for two days (because G is not on the fourth day). So G must be with O:\
- inputs["day3"] = {'O', 'G'}\
- inputs["day4"] = {'J', the person on duty for two days}\
4. K must be scheduled on either the first or second day. We can place K on the first day:\
- inputs["day1"] = {'K', someone}\
5. L and M are left. L can be scheduled on the first day, and M on the second day:\
- inputs["day1"] = {'K', 'L'}\
- inputs["day2"] = {'H', 'M'}\
6. Now, the person on duty for two days can only be M, because everyone else is only on duty once.
Therefore, the final arrangement is:
\begin{verbatim}
{'day1': {'K', 'L'}, 'day2': {'H', 'M'}, 'day3': {'O', 'G'}, 'day4': {'J', 'M'}}
\end{verbatim}
This arrangement satisfies all the given constraints.
\\
\midrule
\textbf{Error Type:} \textbf{Contradictory Conclusion} \\
{\color{blue}There's a contradiction: While step 2 of the reasoning process states that 'J' must be on day 3, the final solution puts it on day 4!}\\
\bottomrule
\end{tabular}
\caption{Error Case: Contradictory Conclusion}
\label{tab:logical-reasoning-example3}
\end{table*}
\endgroup

\begingroup
\begin{table*}[!ht]
\centering
\small
\begin{tabular}{p{\linewidth}}
\toprule
\textbf{\textsc{Prompt (Chinese):}} \begin{CJK*}{UTF8}{gbsn}一个委员会工作两年，每年都由 4 人组成，其中 2 名成员来自下面 4 位法官：F、G、H 和 I，另外 2 名成员来自下面 3 位科学家：V、Y 和 Z。每一年，该委员会有 1 名成员做主席。\
请你生成一个安排方案满足以下约束：\
1. 在第一年做主席的成员在第二年必须退出该委员会。在第二年做主席的人在第一年必须是该委员会的成员。\
2. G 和 V 不能在同一年成为该委员会的成员。\
3. H 和 Y 不能在同一年成为该委员会的成员。\
4. 每一年，I 和 V 中有且只有一位做该委员会的成员。\end{CJK*}\\
\begin{CJK*}{UTF8}{gbsn}
请你一步一步思考，你的安排方案必须按照为以下输入格式要求来进行回答：
输入是一个字典，包含四个键："committee\_year1"、"chairman\_year1"、
"committee\_year2" 和 "chairman\_year2"。
- 在回答的最后，你需要给出一个输入，以表示你的最终安排方案, 其中
`inputs["committee\_year1"]` 是一个包含 4 个元素的 set，包含第一年的委员会
成员的名字。取值范围是`["F","G","H","I","V","Y","Z"]`。
- `inputs["chairman\_year1"]` 是一个字符串，是第一年的主席名字，必须是第
一年的委员。
- `inputs["committee\_year2"]` 是一个包含 4 个元素的 set，包含第二年的委员
会成员的名字。取值范围是 `["F","G","H","I","V","Y","Z"]`。
- `inputs["chairman\_year2"]` 是一个字符串，第二年的主席名字必须是第二年
的委员，并且是第一年的委员。
下面提供一个输入样例（注意这只是一个合法输入的例子，不一定是正确方
案）：
\{'committee\_year1': \{'Z', 'V', 'F', 'G'\}, 'chairman\_year1': 'G', 'committee\_year2': \{'Y',
'V', 'F', 'H'\}, 'chairman\_year2': 'F'\}
\end{CJK*}\\
\midrule
\textbf{\textsc{Prompt (English):}} A committee works for two years, and each year it consists of 4 members, including 2 judges from the following: F, G, H, and I, and 2 scientists from the following: V, Y, and Z. Each year, the committee has one member serve as chairman.\
Please generate an arrangement that satisfies the following constraints:\
1. The member who serves as chairman in the first year must leave the committee in the second year. The member who serves as chairman in the second year must be a member of the committee in the first year.\
2. G and V cannot be members of the committee in the same year.\
3. H and Y cannot be members of the committee in the same year.\
4. Each year, either I or V, but not both, must be a member of the committee.\\
Please think step by step, and your arrangement must follow the input format
requirements below:
The input is a dictionary containing four keys: "committee\_year1",
"chairman\_year1", "committee\_year2", and "chairman\_year2".
- At the end of your answer, you need to provide an input representing your final
arrangement, where inputs["committee\_year1"] is a set containing 4 elements,
representing the members of the committee in the first year. The range of values is
["F","G","H","I","V","Y","Z"].
- inputs["chairman\_year1"] is a string representing the chairman of the first year,
and this person must be a committee member in the first year.
- inputs["committee\_year2"] is a set containing 4 elements, representing the
members of the committee in the second year. The range of values is
["F","G","H","I","V","Y","Z"].
- inputs["chairman\_year2"] is a string representing the chairman of the second
year, and this person must be a committee member in both the current and
previous year.
Below is an example input (note that this is just an example of valid input and may
not be the correct arrangement):
\{'committee\_year1': \{'Z', 'V', 'F', 'G'\}, 'chairman\_year1': 'G', 'committee\_year2': \{'Y',
'V', 'F', 'H'\}, 'chairman\_year2': 'F'\} \\
\midrule
\textbf{\textsc{Model Response (Chinese):}} \begin{CJK*}{UTF8}{gbsn}好的, 我会一步步思考来制定一个满足所有约束条件的安排方案。\
第一步：考虑第一年的委员会成员\
- 需要 2 名法官和 2 名科学家\
- G 和 V 不能同时在委员会中\
- H 和 Y 不能同时在委员会中\
- I 和 V 中必须有且只有一位\
考虑到这些约束，我们可以选择：\
- F, I (法官)\
- V, Z (科学家)\
第二步：选择第一年的主席\
从 F, I, V, Z 中选择一位作为主席。我们选择 F 作为第一年的主席。\
第三步：考虑第二年的委员会成员\
- 第一年的主席 F 必须退出\
- 需要包含第二年主席的人选\
- 仍然需要遵守其他约束条件\
考虑到这些, 我们可以选择：\
- G, H (法官)\
- Y, Z (科学家)\
第四步：选择第二年的主席\
从 G, H, Y, Z 中选择一位作为主席，且这个人必须是第一年的委员。唯一符合条件的是 Z。最终的安排方案如下：\end{CJK*}\\
{\color{darkGreen}
\begin{verbatim}
```python
committee_year1 = {"F", "I", "V", "Z"}
chairman_year1 = "F"
committee_year2 = {"G", "H", "Y", "Z"}
chairman_year2 = "Z"
```
\end{verbatim} \color{blue}{(\begin{CJK*}{UTF8}{gbsn}不符合格式要求\end{CJK*})}}\\
\midrule
\textbf{\textsc{Model Response (English):}} Let's think step by step to create an arrangement that satisfies all the constraints.\

Step 1: Consider the members of the committee for the first year\
- We need 2 judges and 2 scientists\
- G and V cannot both be in the committee\
- H and Y cannot both be in the committee\
- One of I or V must be in the committee, but not both\
Considering these constraints, we can select:\
- Judges: F, I\
- Scientists: V, Z\

Step 2: Select the chairman for the first year\
From F, I, V, Z, select one as chairman. We select F as the chairman for the first year.\

Step 3: Consider the members of the committee for the second year\
- The first year's chairman F must leave the committee\
- The second year's chairman must be included\
- Still need to adhere to the other constraints\
Considering these, we can select:\
- Judges: G, H\
- Scientists: Y, Z\

Step 4: Select the chairman for the second year\
From G, H, Y, Z, select one as chairman, and this person must have been a member in the first year.\
The only eligible person is Z.\
The final arrangement is as follows:
{\color{darkGreen}
\begin{verbatim}
```python
committee_year1 = {"F", "I", "V", "Z"}
chairman_year1 = "F"
committee_year2 = {"G", "H", "Y", "Z"}
chairman_year2 = "Z"
```
\end{verbatim} \color{blue}{(Does not meet the format requirements)}}\\
\midrule
\textbf{Error Type:} \textbf{Inconsistent Format} \\
{\color{blue}The answer uses Python assignment syntax, but a dictionary definition is required.}\\
\bottomrule
\end{tabular}
\caption{Error Case: Inconsistent Format}
\label{tab:logical-reasoning-example4}
\end{table*}
\endgroup

\subsection{Prompt Design}
\label{sec:appendix-prompt}
\subsubsection{Stage 1: Puzzle Formulation}
The first stage focuses on extracting two essential components from the source corpus for open-ended logic puzzle generation: Background information and Logical Constraints. To automate this extraction process, we employ GPT-4o with a one-shot learning approach, providing a single comprehensive example. Table~\ref{tab:appendix-stage1-prompt} details the prompt instructions in both Chinese and English.

\begingroup
\begin{table*}[!ht]
    \centering
    \small
    \begin{tabular}{p{\linewidth}}
        \toprule
        \vspace{-2mm}
        \textbf{\textsc{Instruction\color{darkGreen}{(English)}:}} Provide a description of a problem, which contains some background and definition of the problem (usually the first one or two sentences), as well as some logical constraints. You need to divide these two parts, you can reorganize the language, but do not change the original meaning of the problem. \\
        \vspace{-1mm}
        \textbf{\textsc{Input:}} "A supermarket is neatly arranged with 7 rows of shelves from front to back, displaying 7 categories of goods: stationery, snacks, condiments, daily necessities, alcohol, grains and oils, and beverages, each occupying one row. It is known that: \textbackslash{}nThe alcohol is placed before the condiments; \textbackslash{}nThere are 3 rows between the stationery and the condiments; \textbackslash{}nThe grains and oils are after the snacks, with 2 rows in between; \textbackslash{}nThe daily necessities are either in the row before or after the stationery; \textbackslash{}nThe snacks are in the first row." \\
        \vspace{-2mm}
        \textbf{\textsc{Output:}} 
        \begin{verbatim}
```json
{
    "background": "A supermarket is neatly arranged with 7 rows of shelves from front to back, displaying 7 
    categories of goods: stationery, snacks, condiments, daily necessities, alcohol, grains and oils, 
    and beverages, each occupying one row.",
    "logic_constraints": "The alcohol is placed before the condiments; There are 3 rows between the 
    stationery and the condiments; The grains and oils are after the snacks, with 2 rows in between; The 
    daily necessities are either in the row before or after the stationery; The snacks are in the first row."
}
``` \end{verbatim}\\
    \vspace{-1mm}
    \textbf{\textsc{Input:}} {\color{blue} [[[question]]]}  \\
    \vspace{-2mm}
    \textbf{\textsc{Output:}}  \\
    \midrule
    
    \vspace{-1mm}
    \textbf{\textsc{Instruction\color{darkGreen}{(Chinese)}:}} \begin{CJK*}{UTF8}{gbsn}提供一个问题的描述，这个描述中有一些是问题的背景以及定义(一般就是开头一两句)，也有一些是逻辑的限制条件，你需要把这两部分划分出来，可以自己重新组织语言，但不要改变原题意。\end{CJK*}\\
    \vspace{-1mm}
    \textbf{\textsc{Input:}} \begin{CJK*}{UTF8}{gbsn}"某超市从前到后整齐排列着7排货架，放置着文具、零食、调料、日用品、酒、粮油和饮料7类商品，每类商品\textbackslash{}n占据一排。已知：\textbackslash{}n酒类排在调料类之前；\textbackslash{}n文具类和调料类中间隔着3排；\textbackslash{}n粮油类在零食类之后，中间隔着2排；\textbackslash{}n日用品类紧挨在文具类前一排或者后一排。\textbackslash{}n零食类排在第1排"\end{CJK*}\\
    \vspace{-2mm}
    \textbf{\textsc{Output:}} 
\begin{verbatim}```json 
{
\end{verbatim}
\begin{CJK*}{UTF8}{gbsn}
\qquad"background":"某超市从前到后整齐排列着7排货架，放置着文具、零食、调料、日用品、酒、粮油和饮料7类商品，每类商品占据一排。", \end{CJK*}\\
    \begin{CJK*}{UTF8}{gbsn}
\qquad"logic\_constraints": "酒类排在调料类之前；文具类和调料类中间隔着3排；粮油类在零食类之后，中间隔着2排；日用品类紧挨在文具类前一排或者后一排；零食类排在第1排。"\end{CJK*}
\begin{verbatim}
}
```
\end{verbatim}
\\
    \vspace{-1mm}
    \textbf{\textsc{Input:}} {\color{blue} [[[question]]]}  \\
    \vspace{-2mm}
    \textbf{\textsc{Output:}}  \\ 
    \bottomrule
    \end{tabular}
    \caption{Instructions for Puzzle Formulation}
    \label{tab:appendix-stage1-prompt}
\end{table*}
\endgroup

\subsection{Stage 2: Generation of Format and Verifiers}
\begin{figure*}[ht!]
    \centering
    \includegraphics[width=\linewidth]{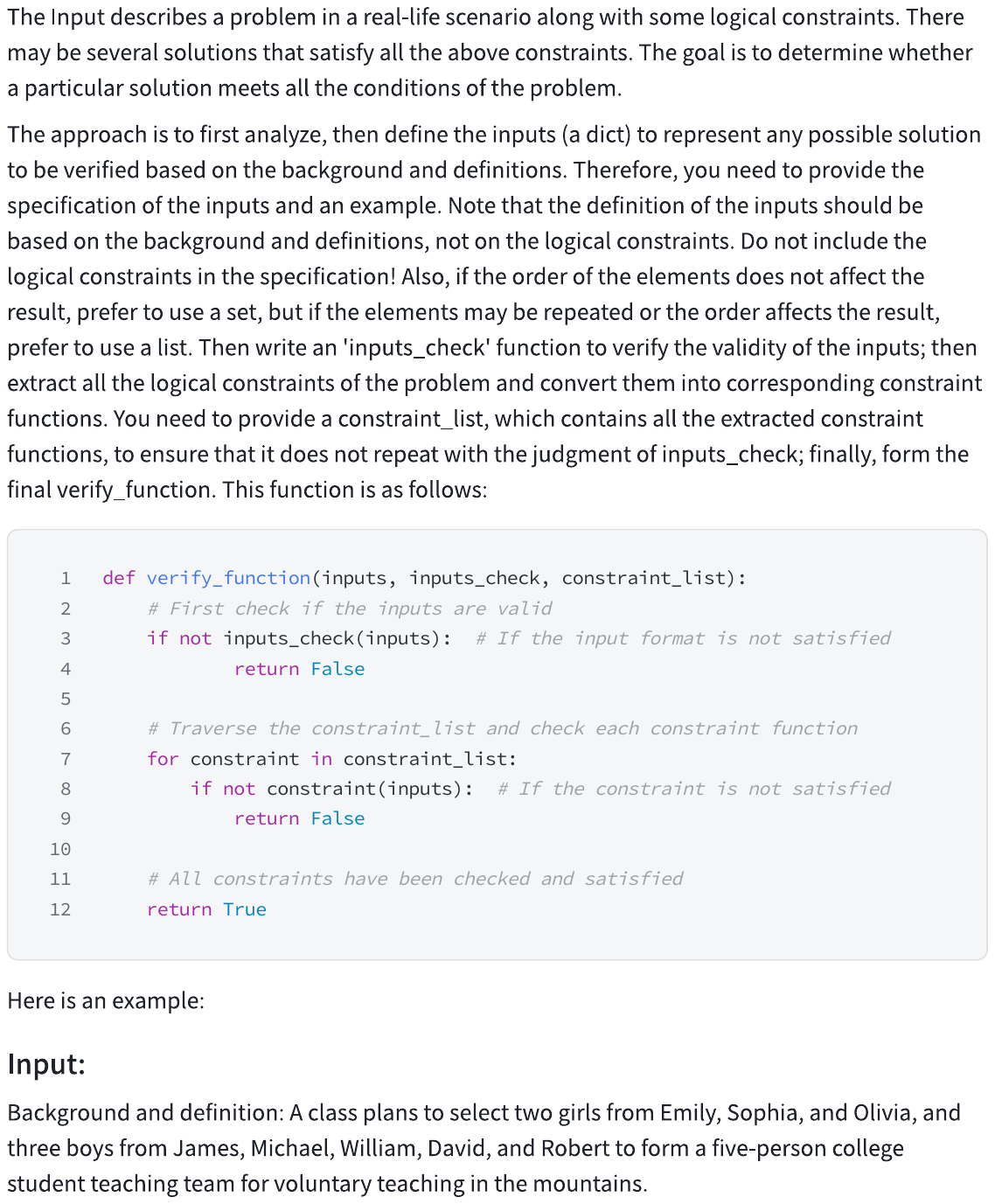}
    \caption{Stage 2 prompt for English Data(1/3). This prompt is designed to simultaneously generate both puzzle requirements (Arrangement Format + Arrangement Example) and program-based verifiers for answer validation, using a one-shot demonstration approach.}
    \label{fig:2-en1}
\end{figure*}

\begin{figure*}[ht!]
    \centering
    \includegraphics[width=\linewidth]{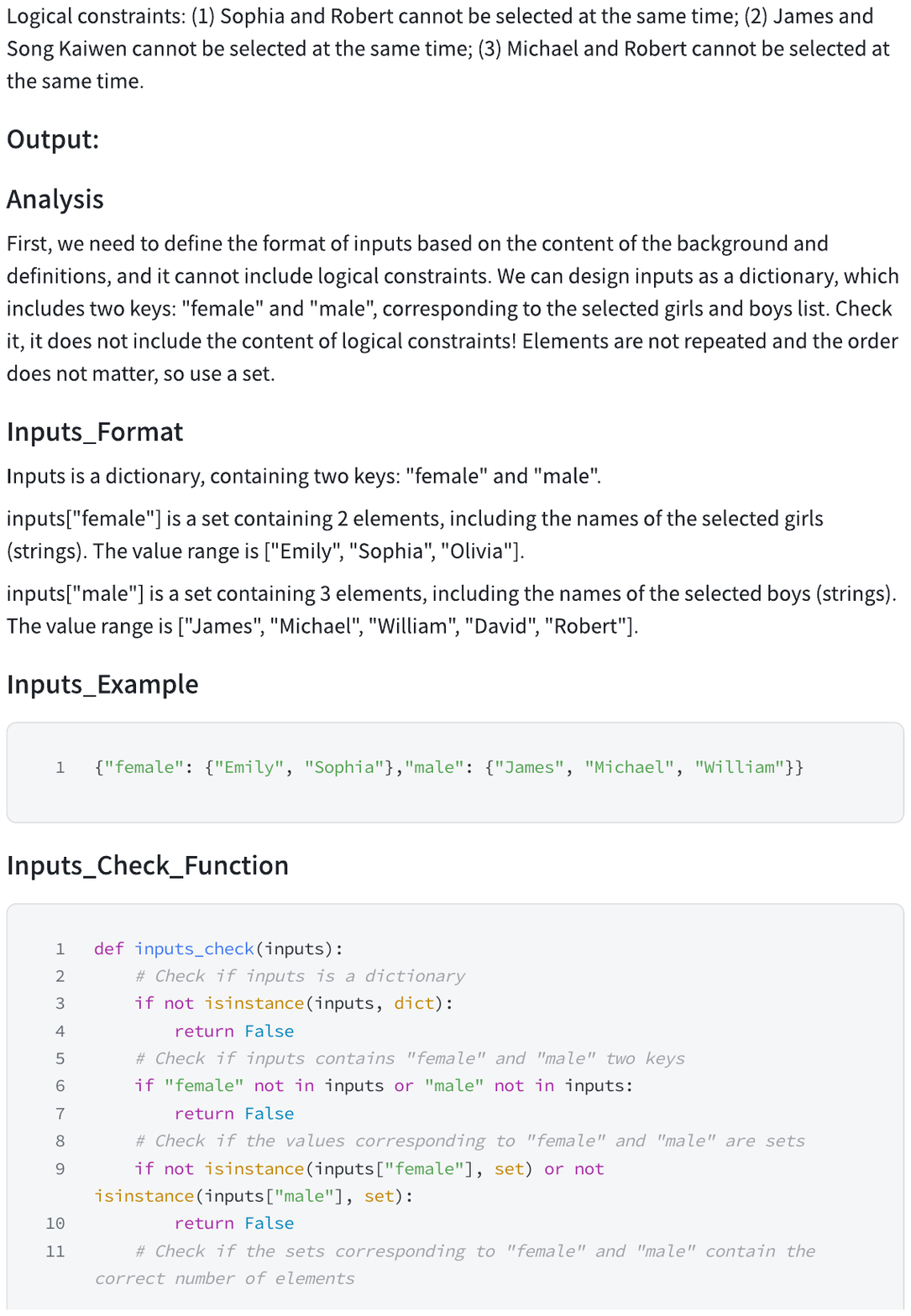}
    \caption{Stage 2 prompt for English Data(2/3). This prompt is designed to simultaneously generate both puzzle requirements (Arrangement Format + Arrangement Example) and program-based verifiers for answer validation, using a one-shot demonstration approach.}
    \label{fig:2-en2}
\end{figure*}

\begin{figure*}[ht!]
    \centering
    \includegraphics[width=\linewidth]{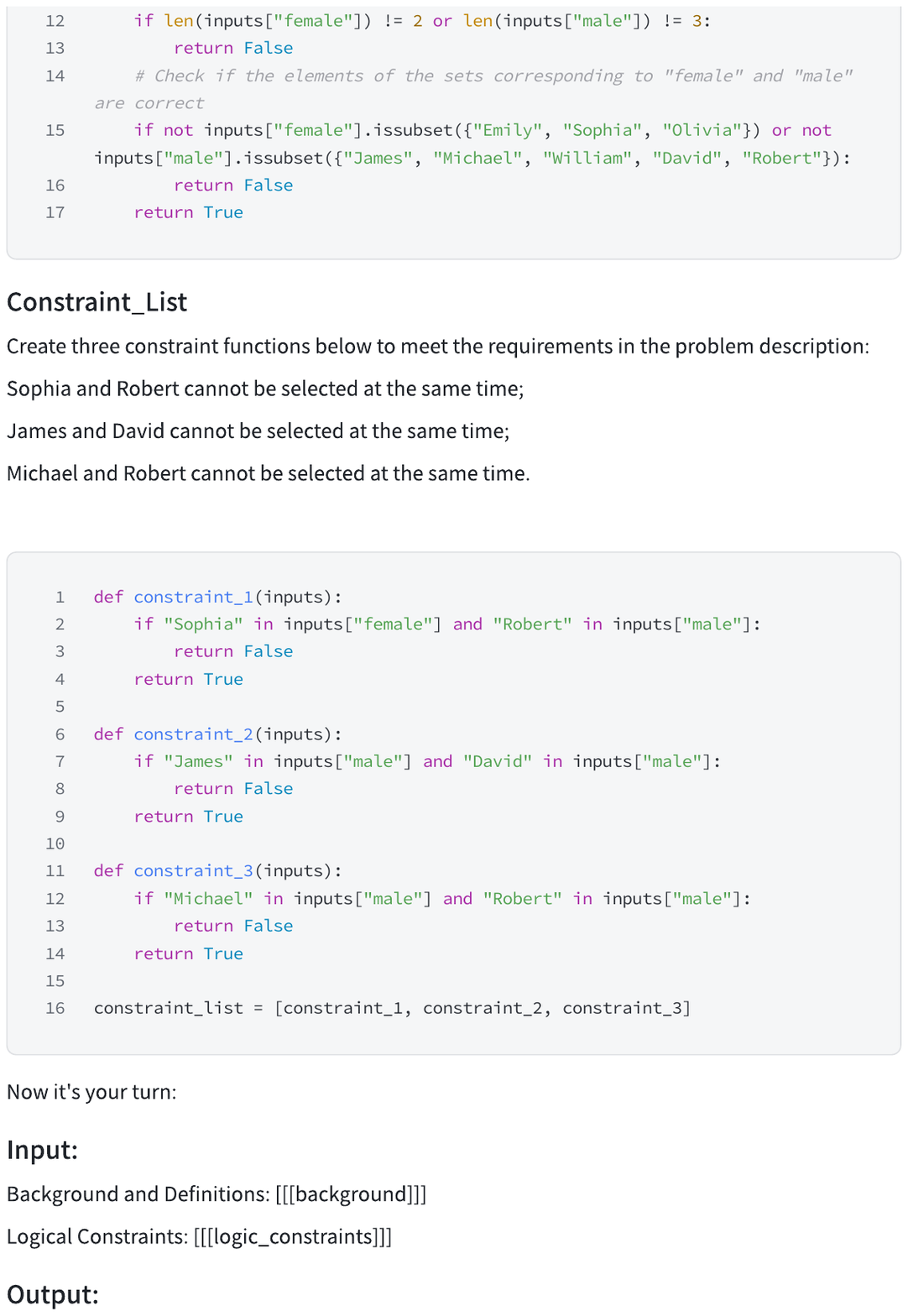}
    \caption{Stage 2 prompt for English Data(3/3). This prompt is designed to simultaneously generate both puzzle requirements (Arrangement Format + Arrangement Example) and program-based verifiers for answer validation, using a one-shot demonstration approach.}
    \label{fig:2-en3}
\end{figure*}

\begin{figure*}[ht!]
    \centering
    \includegraphics[width=\linewidth]{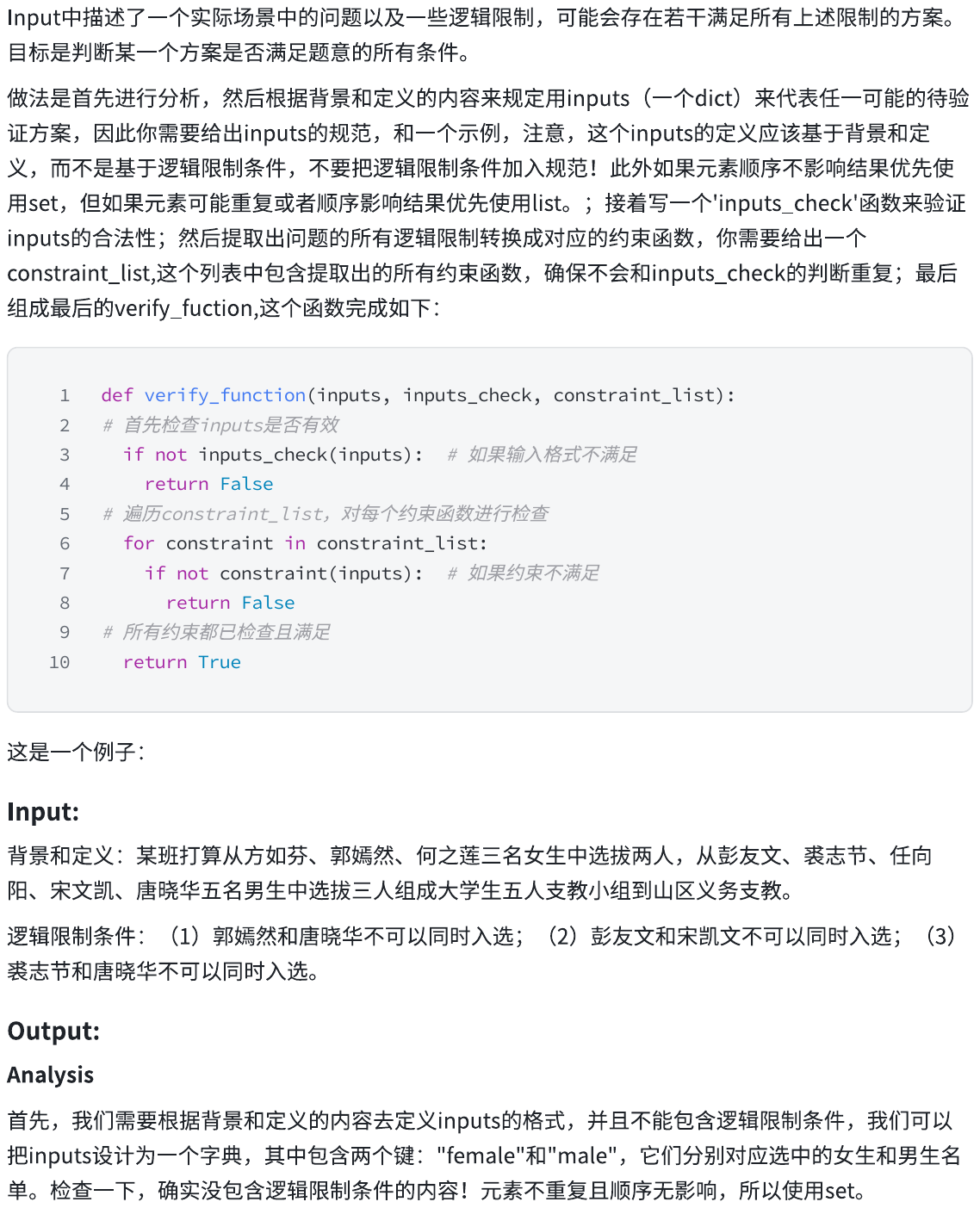}
    \caption{Stage 2 prompt for Chinese Data(1/3). This prompt is designed to simultaneously generate both puzzle requirements (Arrangement Format + Arrangement Example) and program-based verifiers for answer validation, using a one-shot demonstration approach.}
    \label{fig:2-cn1}
\end{figure*}

\begin{figure*}[ht!]
    \centering
    \includegraphics[width=\linewidth]{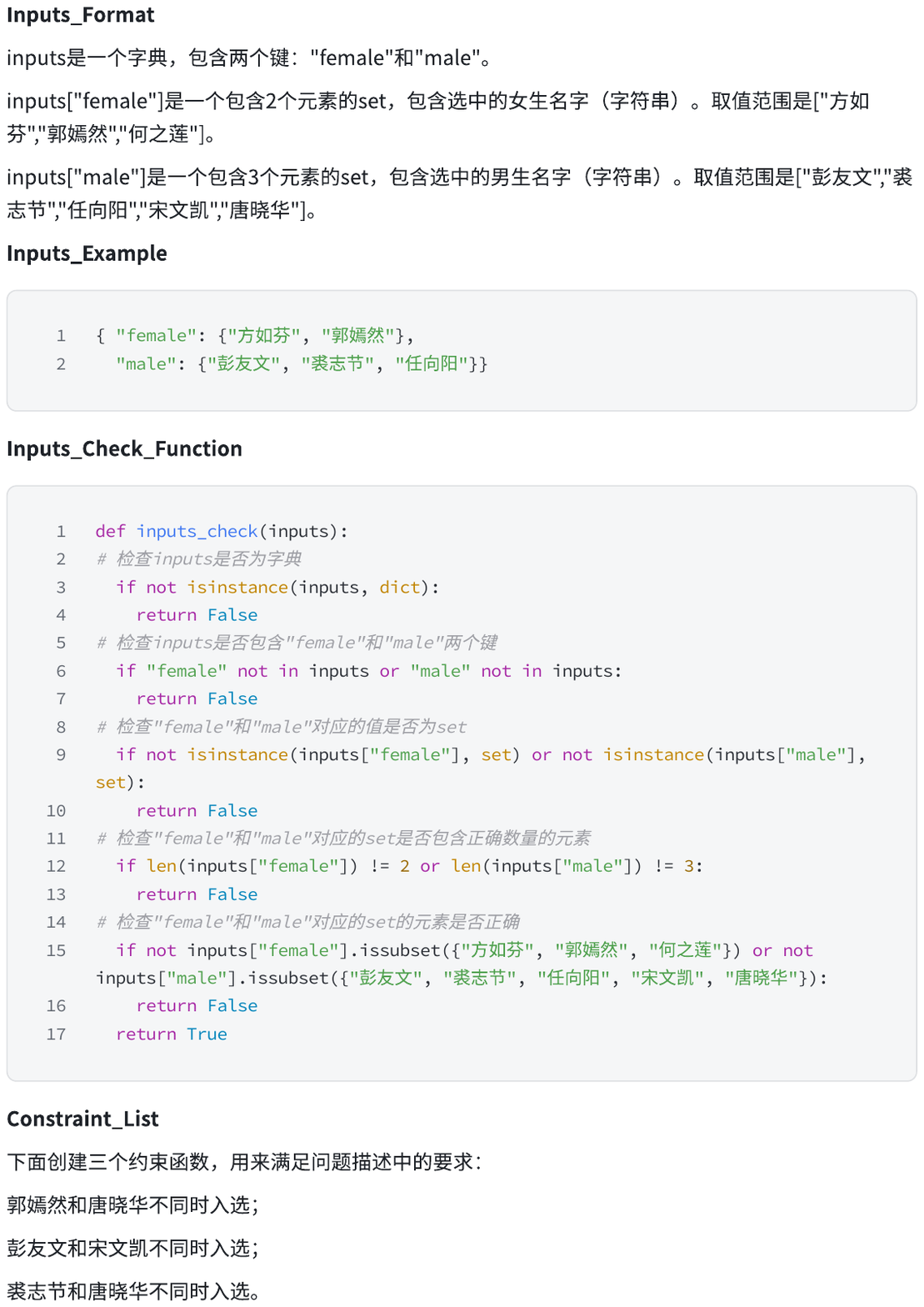}
    \caption{Stage 2 prompt for Chinese Data(2/3). This prompt is designed to simultaneously generate both puzzle requirements (Arrangement Format + Arrangement Example) and program-based verifiers for answer validation, using a one-shot demonstration approach.}
    \label{fig:2-cn2}
\end{figure*}

\begin{figure*}[ht!]
    \centering
    \includegraphics[width=\linewidth]{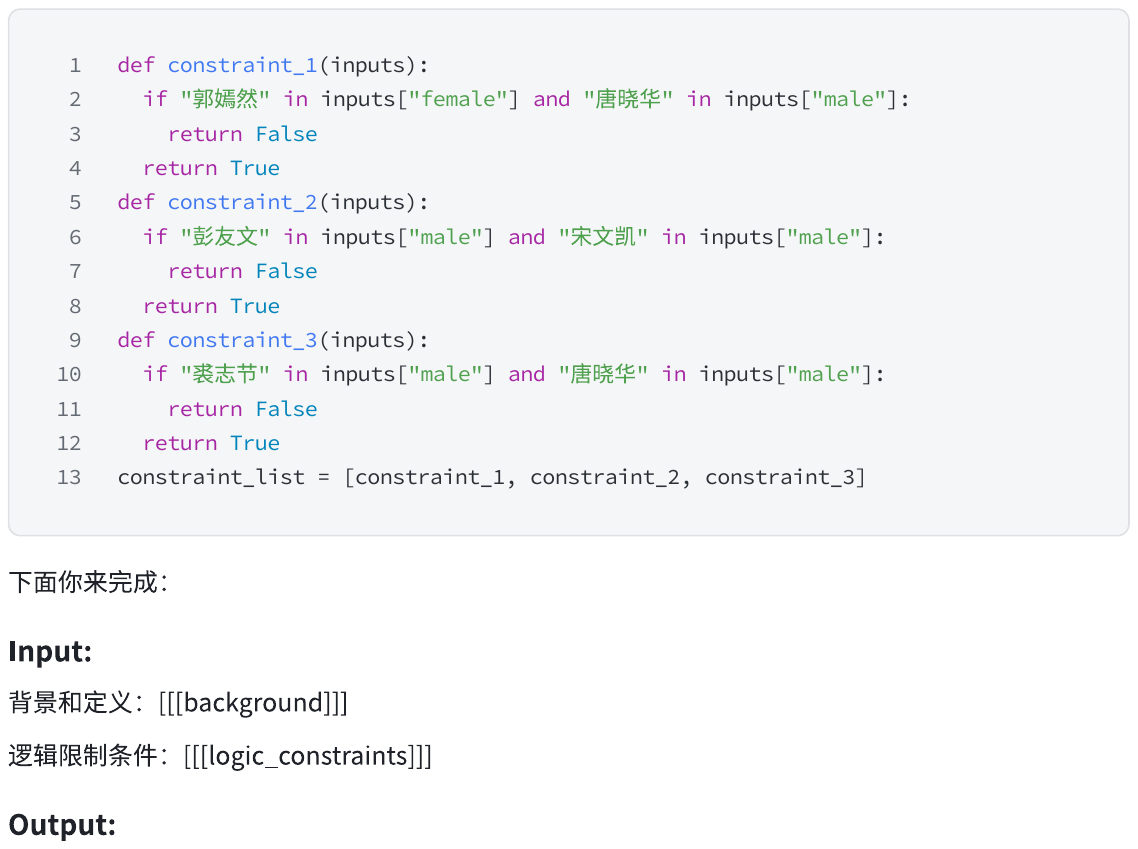}
    \caption{Stage 2 prompt for Chinese Data(3/3).This prompt is designed to simultaneously generate both puzzle requirements (Arrangement Format + Arrangement Example) and program-based verifiers for answer validation, using a one-shot demonstration approach.}
    \label{fig:2-cn3}
\end{figure*}

This stage utilizes prompts to instruct GPT-4o to generate two key components based on the previously extracted Background information and Logical Constraints. The first component is the synthesized puzzle, consisting of the Arrangement Format and an Arrangement Example. The second component comprises two code-based verifiers: a Format Verifier to ensure structural correctness and a Constraint Verifier to validate logical consistency. The detailed prompts in both Chinese and English are presented in Figures~\ref{fig:2-en1}, \ref{fig:2-en2}, \ref{fig:2-en3}, \ref{fig:2-cn1}, \ref{fig:2-cn2}, and \ref{fig:2-cn3}.

\subsection{Stage 2: Validation and Regeneration}
\begin{figure*}[ht!]
    \centering
    \includegraphics[width=\linewidth]{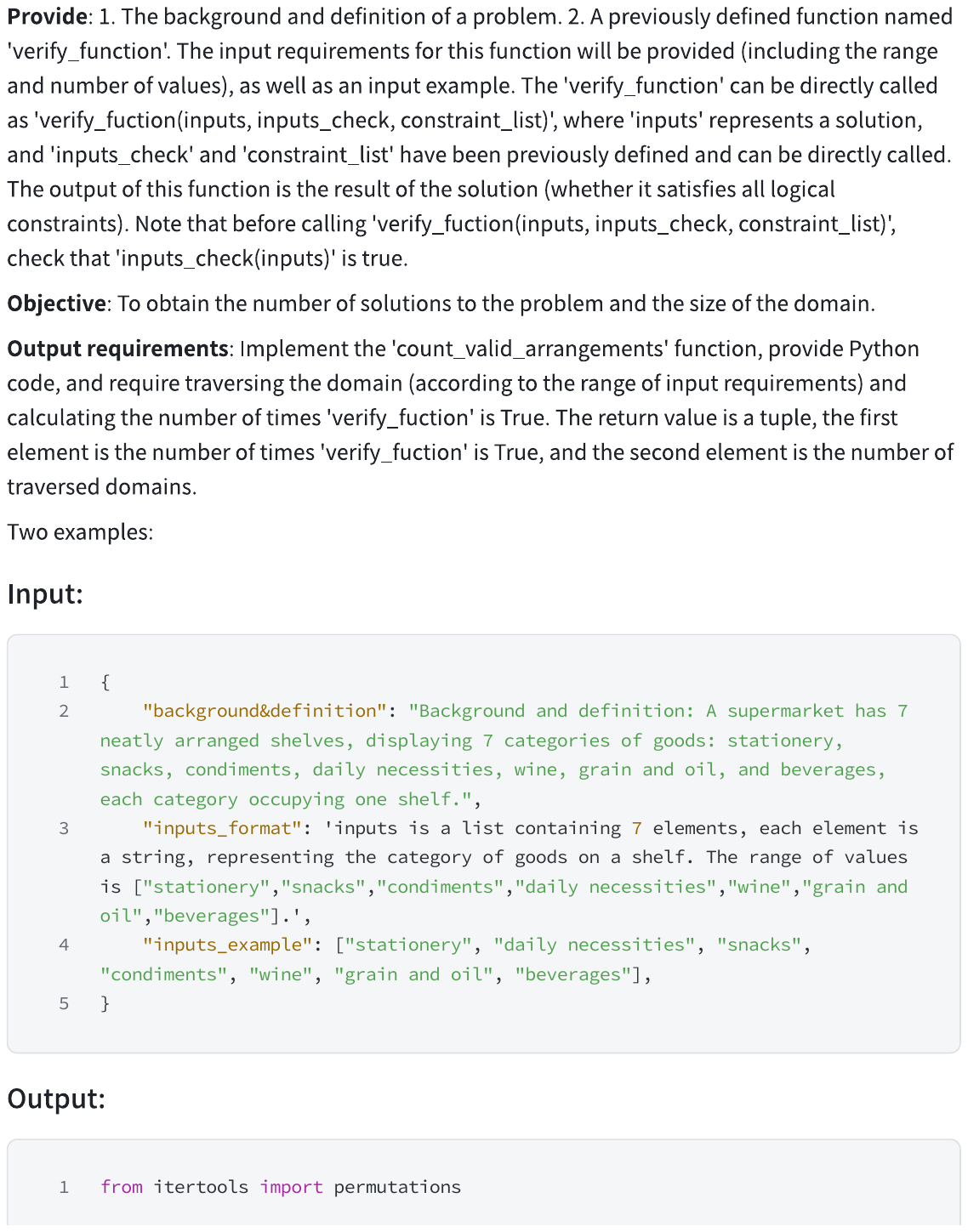}
    \caption{Traversal Function prompt for English Data(1/3). This prompt is designed to generate traversal functions that enumerate all possible solutions, using a two-shots demonstration approach.}
    \label{fig:t-en1}
\end{figure*}

\begin{figure*}[ht!]
    \centering
    \includegraphics[width=\linewidth]{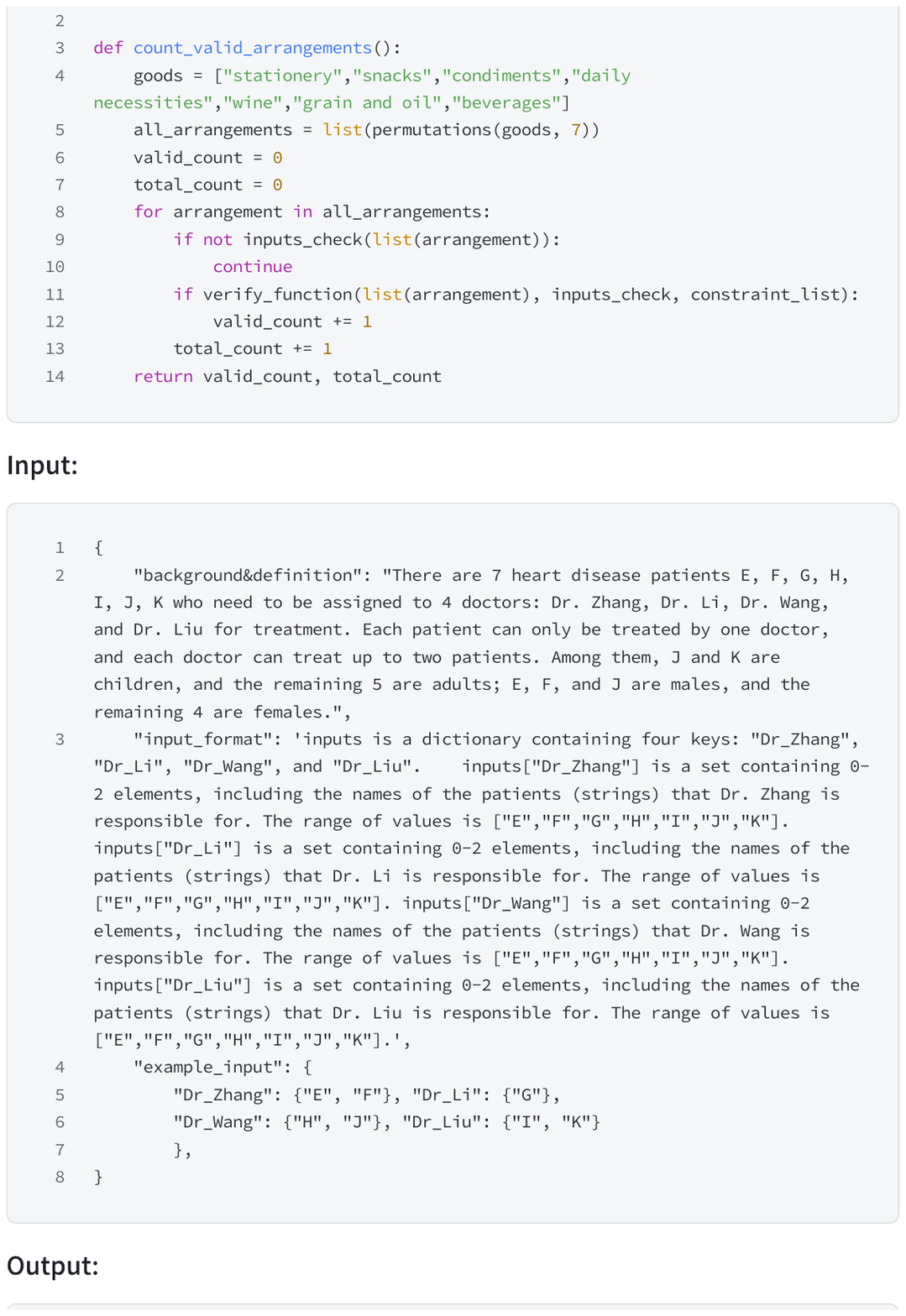}
    \caption{Traversal Function prompt for English Data(2/3). This prompt is designed to generate traversal functions that enumerate all possible solutions, using a two-shots demonstration approach.}
    \label{fig:t-en2}
\end{figure*}

\begin{figure*}[ht!]
    \centering
    \includegraphics[width=\linewidth]{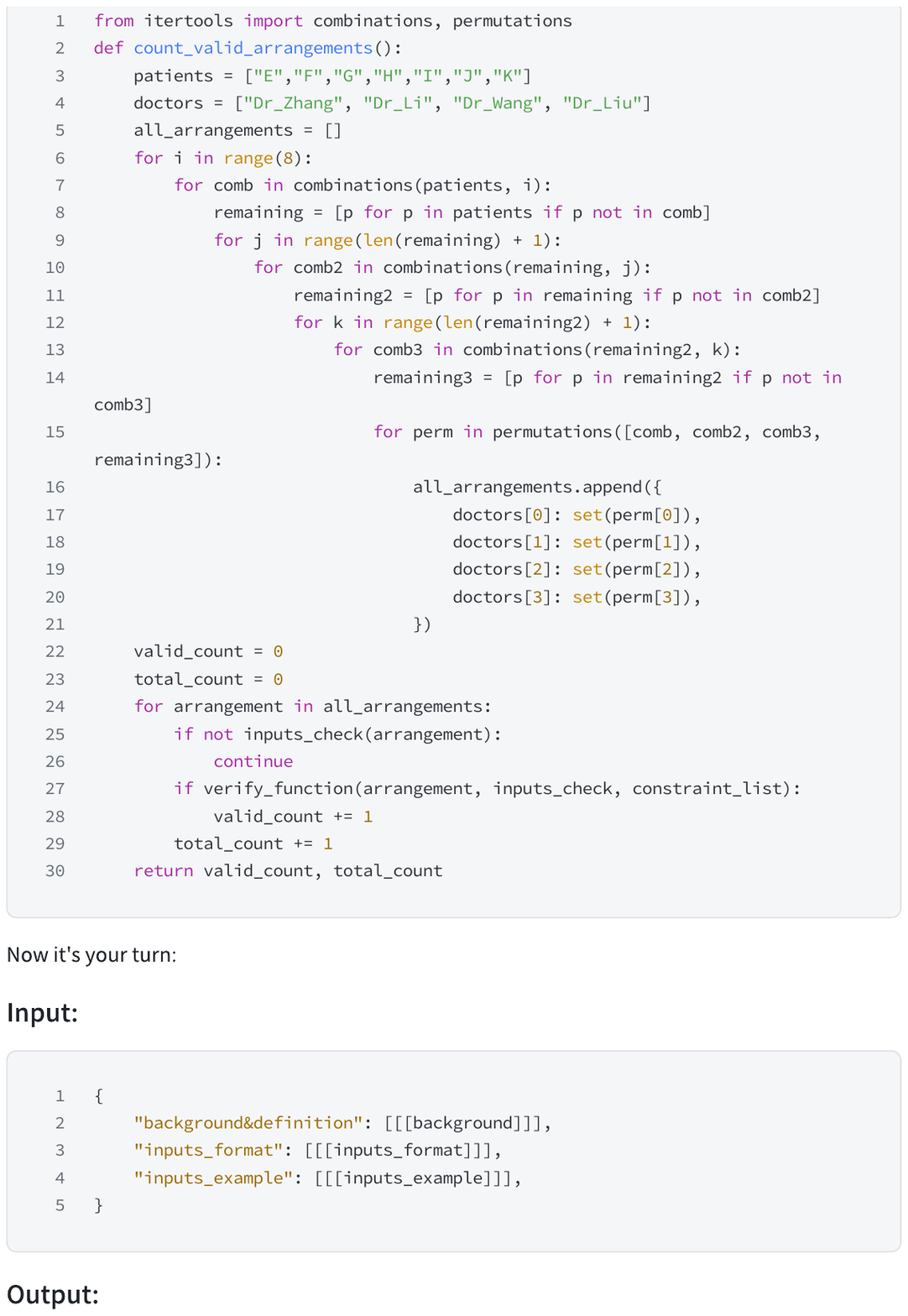}
    \caption{Traversal Function prompt for English Data(3/3). This prompt is designed to generate traversal functions that enumerate all possible solutions, using a two-shots demonstration approach.}
    \label{fig:t-en3}
\end{figure*}

\begin{figure*}[ht!]
    \centering
    \includegraphics[width=\linewidth]{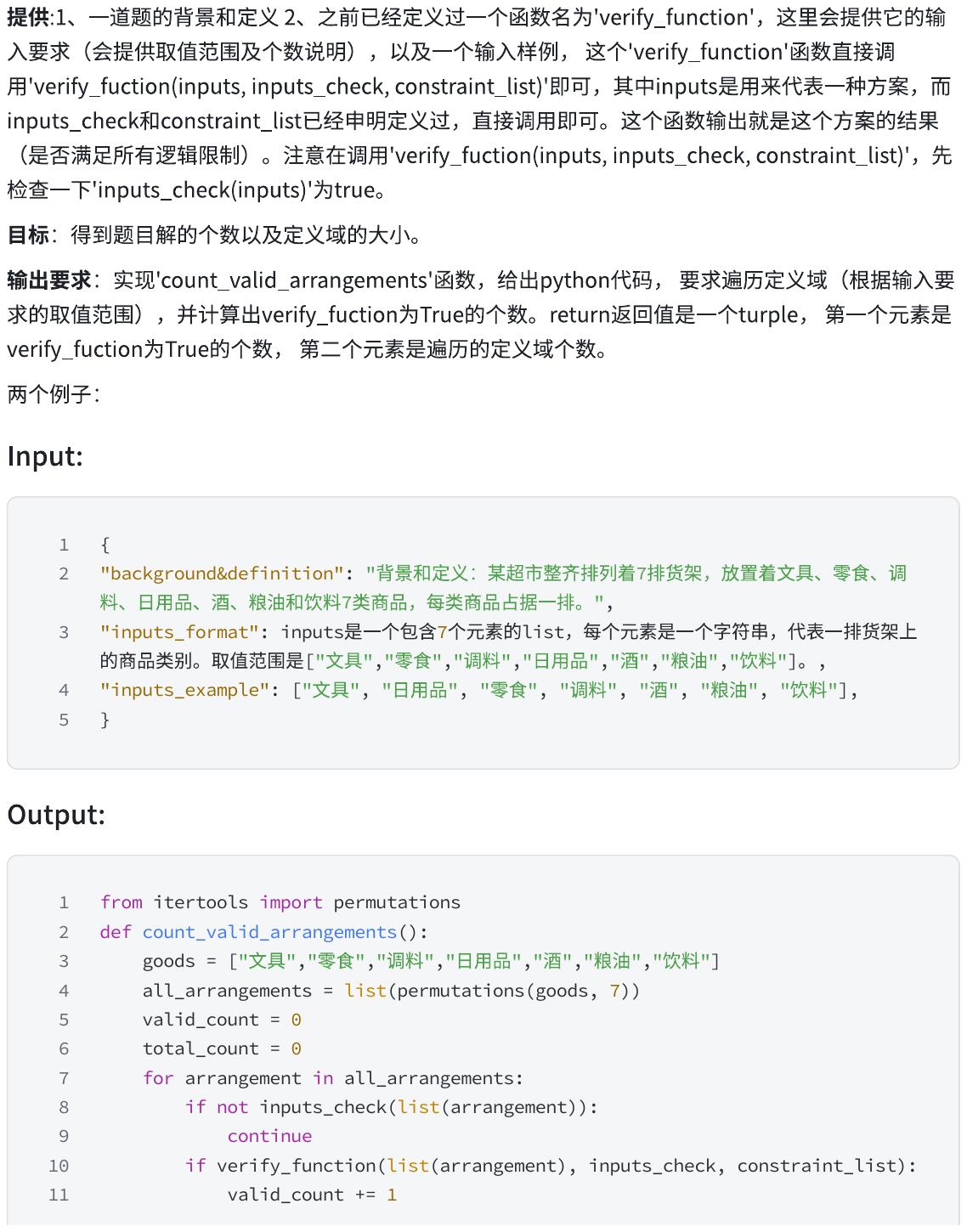}
    \caption{Traversal Function prompt for Chinese Data(1/3). This prompt is designed to generate traversal functions that enumerate all possible solutions, using a two-shots demonstration approach.}
    \label{fig:t-cn1}
\end{figure*}

\begin{figure*}[ht!]
    \centering
    \includegraphics[width=\linewidth]{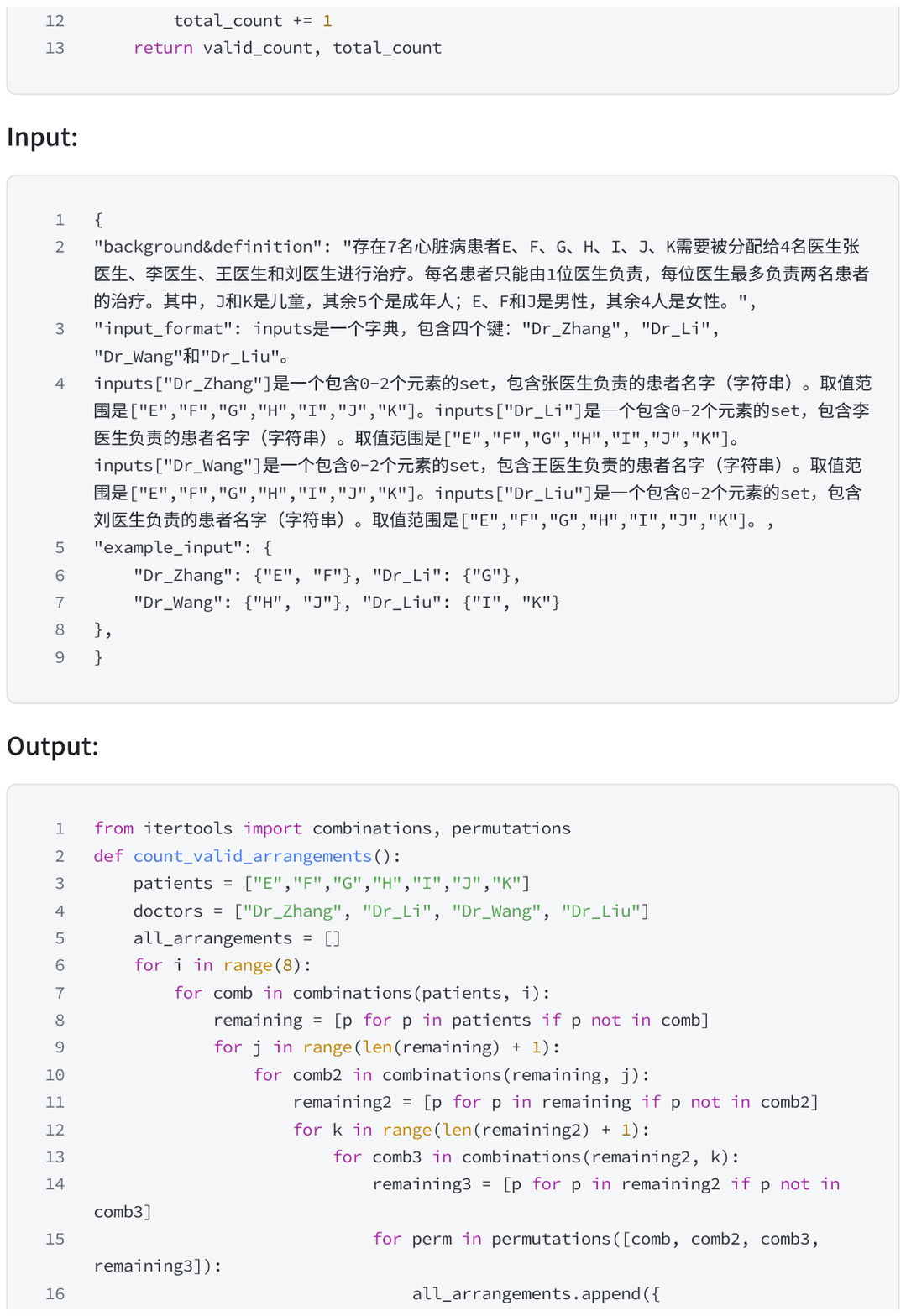}
    \caption{Traversal Function prompt for Chinese Data(2/3). This prompt is designed to generate traversal functions that enumerate all possible solutions, using a two-shots demonstration approach.}
    \label{fig:t-cn2}
\end{figure*}

\begin{figure*}[ht!]
    \centering
    \includegraphics[width=\linewidth]{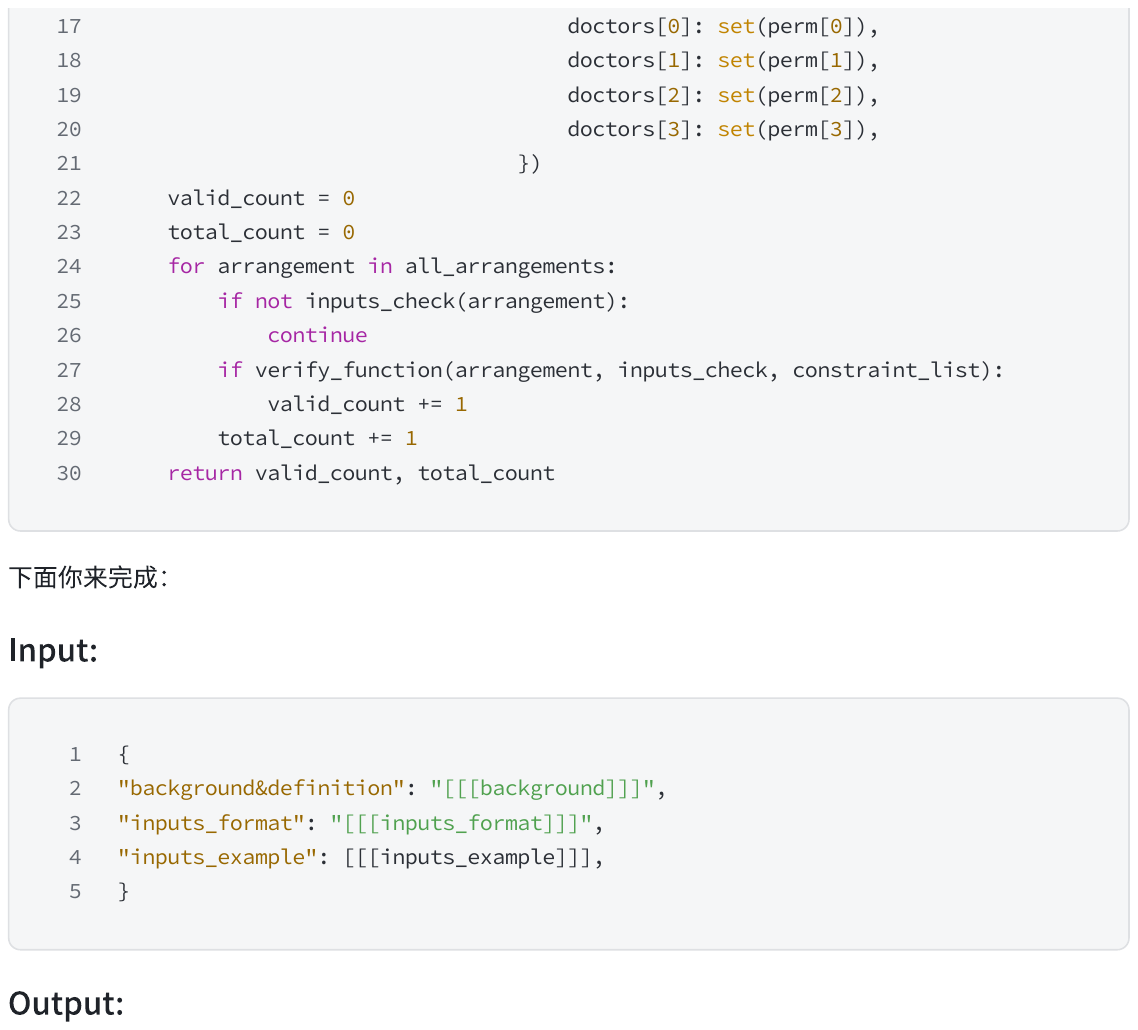}
    \caption{Traversal Function prompt for Chinese Data(3/3). This prompt is designed to generate traversal functions that enumerate all possible solutions, using a two-shots demonstration approach.}
    \label{fig:t-cn3}
\end{figure*}

A crucial verification step involves the implementation of a Traversal Function to validate the correctness of the generated puzzles. This function employs an enumeration approach to verify the existence of solutions that satisfy both verifiers established in the previous stage. By exhaustively exploring the value ranges of puzzle parameters, the Traversal Function helps ensure the quality and solvability of the synthesized puzzles. We utilize a two-shot prompting approach to guide GPT-4o in generating this function based on the value domains specified in the puzzle. The detailed prompts in both English and Chinese can be found in Figures~\ref{fig:t-en1}, \ref{fig:t-en2}, \ref{fig:t-en3}, \ref{fig:t-cn1}, \ref{fig:t-cn2}, and \ref{fig:t-cn3}.

\subsection{Stage 3: Data Augmentation}
\begin{figure*}[ht!]
    \centering
    \includegraphics[width=\linewidth]{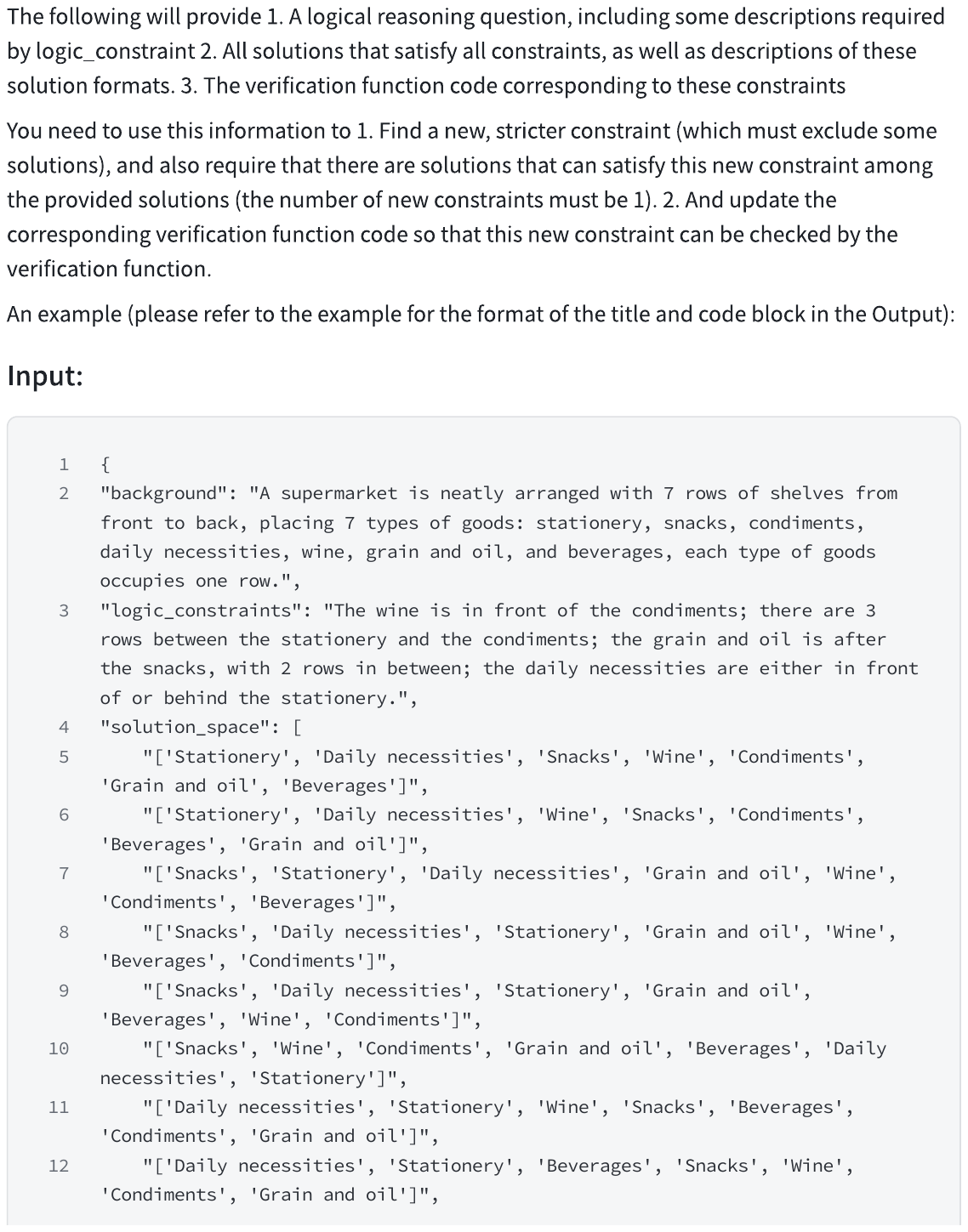}
    \caption{Stage 3 prompt for English Data(1/3). This prompt is designed to augment logical constraints by generating both textual descriptions and corresponding program-based verifiers, using a one-shot demonstration approach.}
    \label{fig:3-en1}
\end{figure*}

\begin{figure*}[ht!]
    \centering
    \includegraphics[width=\linewidth]{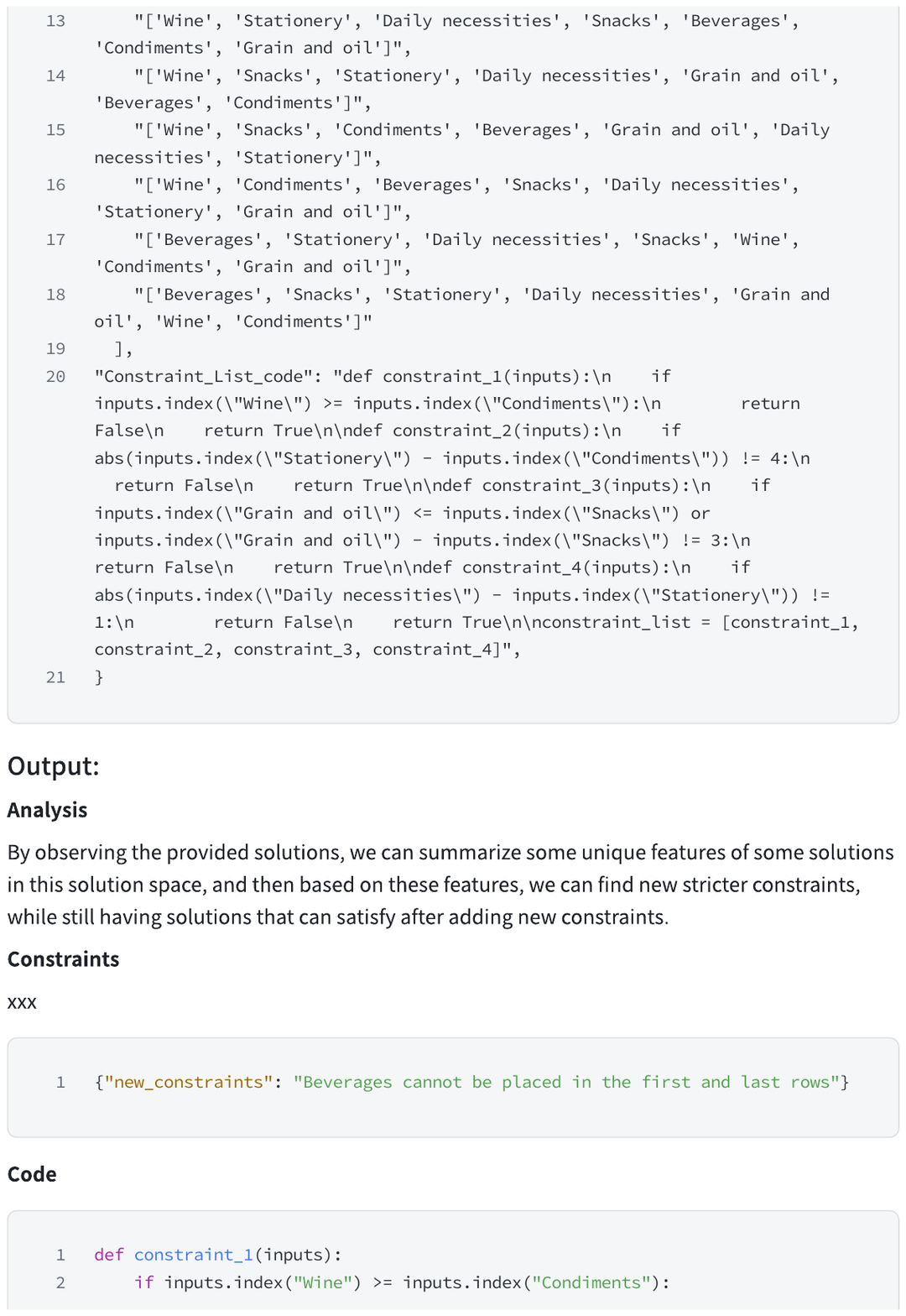}
    \caption{Stage 3 prompt for English Data(2/3). This prompt is designed to augment logical constraints by generating both textual descriptions and corresponding program-based verifiers, using a one-shot demonstration approach.}
    \label{fig:3-en2}
\end{figure*}

\begin{figure*}[ht!]
    \centering
    \includegraphics[width=\linewidth]{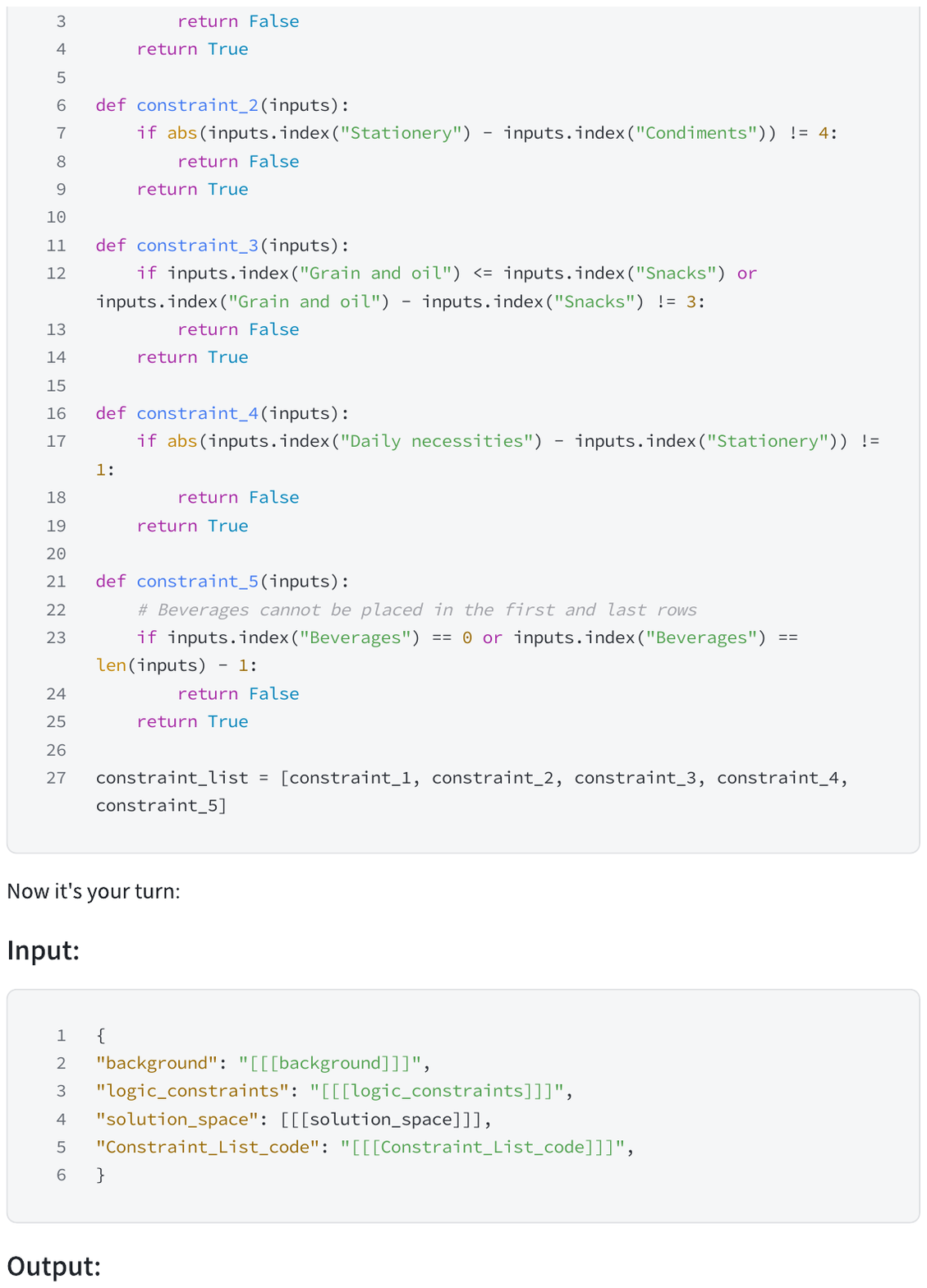}
    \caption{Stage 3 prompt for English Data(3/3). This prompt is designed to augment logical constraints by generating both textual descriptions and corresponding program-based verifiers, using a one-shot demonstration approach.}
    \label{fig:3-en3}
\end{figure*}

\begin{figure*}[ht!]
    \centering
    \includegraphics[width=\linewidth]{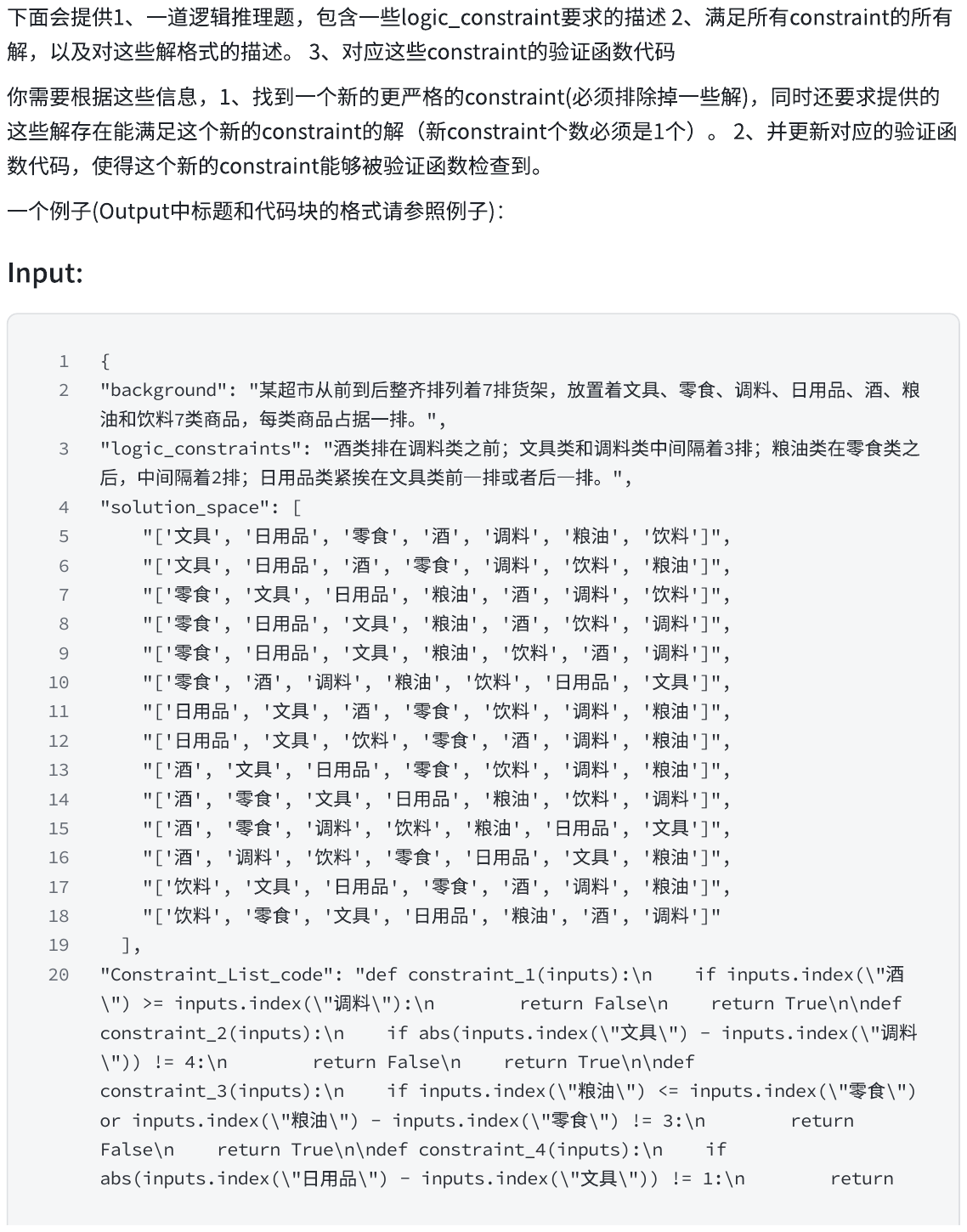}
    \caption{Stage 3 prompt for Chinese Data(1/3). This prompt is designed to augment logical constraints by generating both textual descriptions and corresponding program-based verifiers, using a one-shot demonstration approach.}
    \label{fig:3-cn1}
\end{figure*}

\begin{figure*}[ht!]
    \centering
    \includegraphics[width=\linewidth]{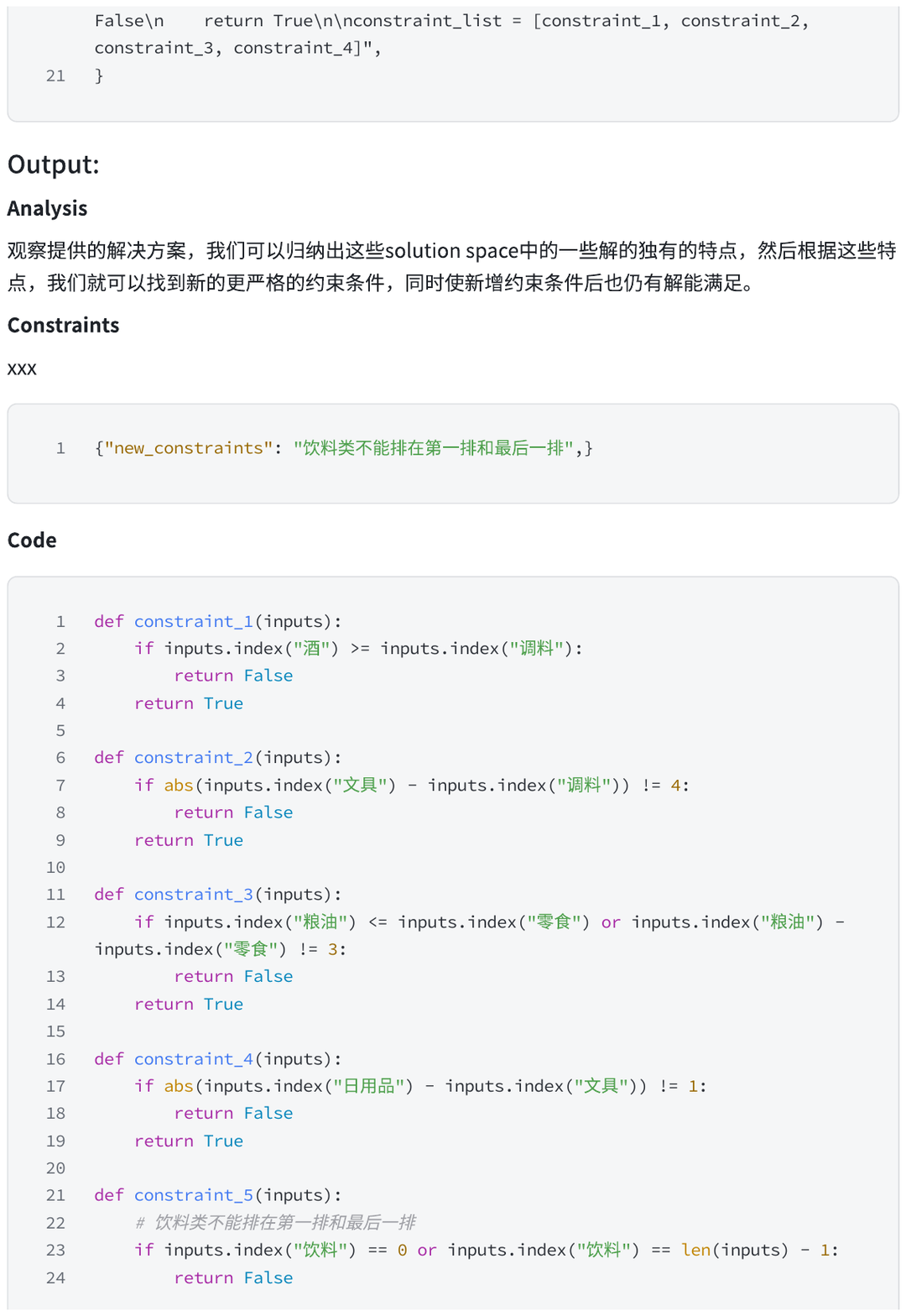}
    \caption{Stage 3 prompt for Chinese Data(2/3). This prompt is designed to augment logical constraints by generating both textual descriptions and corresponding program-based verifiers, using a one-shot demonstration approach.}
    \label{fig:3-cn2}
\end{figure*}

\begin{figure*}[ht!]
    \centering
    \includegraphics[width=\linewidth]{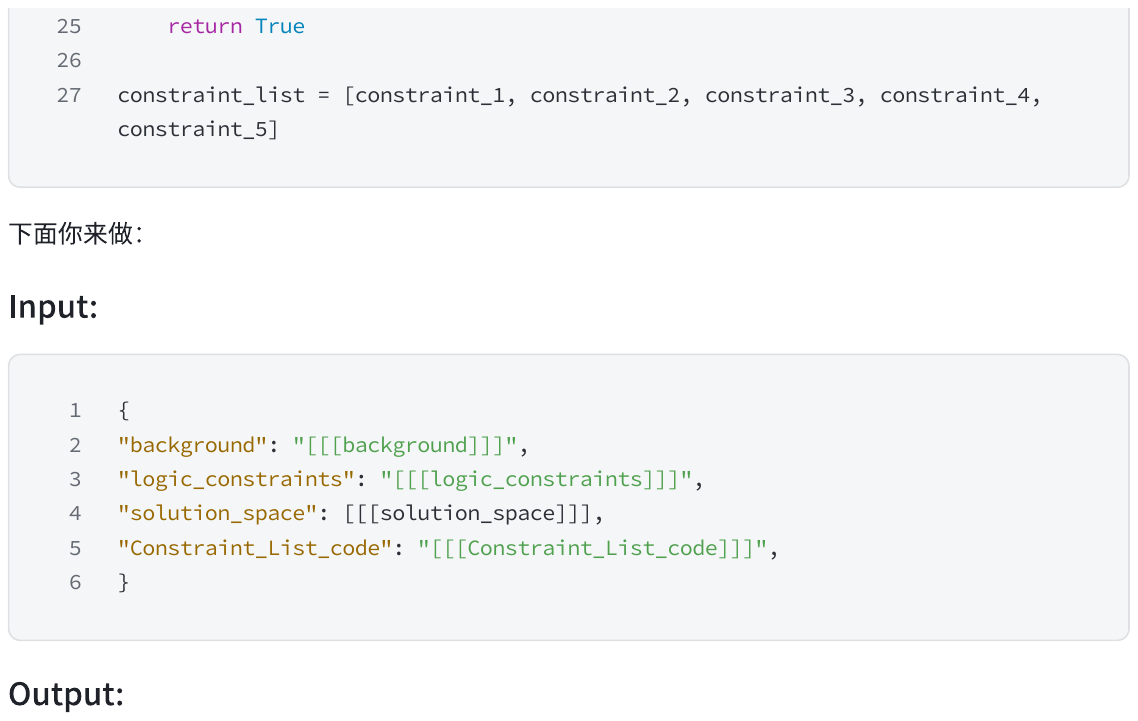}
    \caption{Stage 3 prompt for Chinese Data(3/3). This prompt is designed to augment logical constraints by generating both textual descriptions and corresponding program-based verifiers, using a one-shot demonstration approach.}
    \label{fig:3-cn3}
\end{figure*}

The third stage implements data augmentation through two approaches: Reduction and Expansion. In the Expansion process, we leverage GPT-4o to generate additional constraints by learning from existing ones and their known solution sets. The model analyzes the patterns in current constraints and their corresponding solutions to synthesize new, compatible constraints. To ensure the validity of the augmented puzzles, we employ the Traversal Function for verification, confirming that each expanded puzzle maintains solvability. The complete prompts for this stage are presented in Figures~\ref{fig:3-en1}, \ref{fig:3-en2}, \ref{fig:3-en3}, \ref{fig:3-cn1}, \ref{fig:3-cn2}, and \ref{fig:3-cn3}.




\subsection{Detailed Training Results}
\label{sec:appendix-training-details}
The experimental results presented in Table~\ref{tab:training-set-results-appendix} report the accuracy scores with their corresponding standard deviations (std), calculated across five independent trials.

\begin{table*}[!ht]
\centering
\small
\begin{tabular}{lcccccc}
\toprule
\multirow{2}{*}{\textbf{Model}} & \multicolumn{2}{c}{\textbf{AutoLogi}} & \multirow{2}{*}{\textbf{AR-LSAT}} & \multirow{2}{*}{\textbf{LogiQA}} & \multirow{2}{*}{\textbf{MUSR}} & \multirow{2}{*}{\textbf{LiveBench}} \\ \cmidrule(lr){2-3}
& \textbf{EN} & \textbf{CN} & & & & \\
\midrule
\textit{Baseline} Qwen2.5-7b-instruct & $\text{43.64}{\textcolor{deepgreen}{\pm\text{1.25}}}$ & $\text{42.08}{\textcolor{deepgreen}{\pm\text{1.50}}}$ & $\text{22.70}{\textcolor{deepgreen}{\pm\text{0.84}}}$ & $\text{34.42}{\textcolor{deepgreen}{\pm\text{1.38}}}$ & $\text{47.14}{\textcolor{deepgreen}{\pm\text{0.73}}}$ & $\text{30.67}^{\dagger}$ \\
+\textit{Self-Alignment}  $DPO$ & $\text{48.80}{\textcolor{deepgreen}{\pm\text{0.91}}}$ & $\text{45.39}{\textcolor{deepgreen}{\pm\text{0.41}}}$ & $\text{26.09}{\textcolor{deepgreen}{\pm\text{2.37}}}$ & $\text{38.05}{\textcolor{deepgreen}{\pm\text{4.00}}}$ & $\text{47.57}{\textcolor{deepgreen}{\pm\text{0.30}}}$ & $\text{35.73}{\textcolor{deepgreen}{\pm\text{2.94}}}$ \\
+\textit{Strong-to-Weak} $RFT$ & $\text{48.33}{\textcolor{deepgreen}{\pm\text{0.63}}}$ & $\text{47.54}{\textcolor{deepgreen}{\pm\text{0.71}}}$ & $\text{27.30}{\textcolor{deepgreen}{\pm\text{0.97}}}$ & $\text{35.93}{\textcolor{deepgreen}{\pm\text{2.59}}}$ & $\text{49.15}{\textcolor{deepgreen}{\pm\text{0.55}}}$ & $\text{30.07}{\textcolor{deepgreen}{\pm\text{2.00}}}$ \\
\midrule
\textit{Baseline} Qwen2.5-72b-instruct & $\text{68.18}{\textcolor{deepgreen}{\pm\text{0.77}}}$ & $\text{63.92}{\textcolor{deepgreen}{\pm\text{0.56}}}$ & $\text{31.65}{\textcolor{deepgreen}{\pm\text{1.89}}}$ & $\text{47.92}{\textcolor{deepgreen}{\pm\text{1.37}}}$ & $\text{54.21}{\textcolor{deepgreen}{\pm\text{0.41}}}$ & $\text{46.00}^{\dagger}$ \\
+\textit{Self-Alignment} $DPO$ & $\text{74.79}{\textcolor{deepgreen}{\pm\text{0.41}}}$ & $\text{69.54}{\textcolor{deepgreen}{\pm\text{0.87}}}$ & $\text{38.70}{\textcolor{deepgreen}{\pm\text{1.98}}}$ & $\text{48.14}{\textcolor{deepgreen}{\pm\text{3.98}}}$ & $\text{56.48}{\textcolor{deepgreen}{\pm\text{0.47}}}$ & $\text{52.13}{\textcolor{deepgreen}{\pm\text{1.48}}}$ \\
\bottomrule
\end{tabular}
\caption{Results on various benchmarks. Numbers in parentheses indicate standard deviation over 5 trials. $^{\dagger}$Results reported from LiveBench leaderboard(2024-08-31).}
\label{tab:training-set-results-appendix}
\end{table*}